\documentclass{article}

\usepackage{microtype}
\usepackage{graphicx}
\usepackage{subfigure}
\usepackage{booktabs} % for professional tables

\usepackage[utf8]{inputenc} % allow utf-8 input
\usepackage[T1]{fontenc}    % use 8-bit T1 fonts
\usepackage{xcolor}
\usepackage{soul}
\usepackage{xspace}
\usepackage{enumitem}
\usepackage{balance}
\usepackage{multirow}
\usepackage{booktabs}
\usepackage{siunitx}
\usepackage{float}
\usepackage{tabularx}
\usepackage{array}
\usepackage{makecell}
\usepackage[ruled,vlined]{algorithm2e}

\setlist[enumerate]{itemsep=2pt, parsep=0pt}
\definecolor{darkblue}{rgb}{0,0.08,0.45}
\definecolor{custompink}{RGB}{237,3,140}  
\usepackage[most]{tcolorbox}

\usepackage{hyperref}

\newcommand{\fanlang}{\textit{FanLang-9}\xspace}
\newcommand{\Listener}{\textit{Listener}\xspace}
\newcommand{\Thinker}{\textit{Thinker}\xspace}
\newcommand{\Presenter}{\textit{Presenter}\xspace}

\usepackage[accepted]{icml2024}

\usepackage{amsmath}
\usepackage{amssymb}
\usepackage{mathtools}
\usepackage{amsthm}

\usepackage{siunitx}
\definecolor{lightgray}{gray}{0.98}
\usepackage{listings}
\usepackage{caption}
\usepackage{graphicx}

\usepackage[capitalize,noabbrev]{cleveref}

\theoremstyle{plain}

\theoremstyle{definition}

\theoremstyle{remark}

\usepackage[textsize=tiny]{todonotes}

\icmltitlerunning{Enhance Reasoning for Large Language Model in the Game Werewolf}

\begin{document}

\twocolumn[
\icmltitle{Enhance Reasoning for Large Language Models in the Game Werewolf}



\icmlsetsymbol{equal}{*}

\begin{icmlauthorlist}
\icmlauthor{Shuang Wu}{affiliation1,equal}
\icmlauthor{Liwen Zhu}{affiliation1,equal}
\icmlauthor{Tao Yang}{affiliation1}
\icmlauthor{Shiwei Xu}{affiliation1}
\icmlauthor{Qiang Fu}{affiliation1}
\icmlauthor{Wei Yang}{affiliation1}
\icmlauthor{Haobo Fu}{affiliation1}
\end{icmlauthorlist}

\icmlaffiliation{affiliation1}{Tencent AI Lab, Shenzhen, China}
\icmlcorrespondingauthor{Shuang Wu}{shawnswu@tencent.com}

\icmlkeywords{Machine Learning, ICML}

\vskip 0.3in
]



\printAffiliationsAndNotice{\icmlEqualContribution} % otherwise use the standard text.

\begin{abstract}
This paper presents an innovative framework that integrates Large Language Models (LLMs) with an external \Thinker module to enhance the reasoning capabilities of LLM-based agents.
Unlike augmenting LLMs with prompt engineering, \Thinker directly harnesses knowledge from databases and employs various optimization techniques.
The framework forms a reasoning hierarchy where LLMs handle intuitive \emph{System-1} tasks such as natural language processing, 
while the \Thinker focuses on cognitive \emph{System-2} tasks that require complex logical analysis and domain-specific knowledge.
Our framework is presented using a 9-player Werewolf game that demands dual-system reasoning.
We introduce a communication protocol between LLMs and the \Thinker, and train the \Thinker using data from \num{18800} human sessions and reinforcement learning.
Experiments demonstrate the framework's effectiveness in deductive reasoning, speech generation, and online game evaluation.
Additionally, we fine-tune a 6B LLM to surpass GPT4 when integrated with the \Thinker.
This paper also contributes the largest dataset \href{https://github.com/boluoweifenda/werewolf}{\color{custompink}https://github.com/boluoweifenda/werewolf} for social deduction games to date.  
\end{abstract}

\section{Introduction}
\label{introduction}
The field of artificial intelligence has witnessed groundbreaking advancements in recent years, with the development of Large Language Models (LLMs)~\cite{ouyang2022training,openai2023gpt,anil2023palm}.
Apart from their impressive proficiency in natural language processing (NLP) tasks~\cite{thoppilan2022lamda,zhang2023benchmarking}, LLMs also exhibit vast potential as a general problem solver in areas such as planning and decision-making~\cite{huang2022language}, knowledge transfer and generalization~\cite{anil2022exploring} and multi-modal perception~\cite{yin2023survey} due to the rich world knowledge embedded in their training corpora.
As a result, integrating LLMs as central controllers with task agents to enable end-to-end solutions has become one of the most promising research directions, leading to significant breakthroughs in domains such as tools and assistants~\cite{schick2023toolformer,ge2023openagi}, engineering~\cite{ahn2022can}, social simulations~\cite{park2023generative}, and gaming~\cite{wang2023voyager}.

LLM-based agents harness LLMs for their general-purpose reasoning abilities~\cite{huang2022towards}, which are primarily enabled by prompt engineering methods such as
information profiling~\cite{zhang2023building,qian2023communicative},
step-by-step task decomposition~\cite{wei2022chain,zhou2022least}, recursive prompting by feedback from the environment~\cite{yao2022react}, human~\cite{wu2022ai} and self-refinement~\cite{madaan2023self,shinn2023reflexion}.
These methods thus eliminate the requirement for domain-specific fine-tuning of LLMs.
To augment their task-specific competencies, researchers also adopt external modules such as memory for storing and retrieving historical information~\cite{lin2023agentsims,zhong2023memorybank,hu2023chatdb}, external tools~\cite{schick2023toolformer}, APIs~\cite{qin2023toolllm}, knowledge bases~\cite{lewis2020retrieval} and expert models~\cite{yang2023mm,ge2023openagi}.

Despite these advancements, challenges persist in domain-specific tasks, where LLM-based agents often serve primarily as demonstrations rather than as practical solutions~\cite{qian2023communicative,liu2023agentbench}. 
First, while LLMs have emerged some basic reasoning capabilities, they require sufficient model scales~\cite{kaplan2020scaling} and substantial computational overheads, along with various aforementioned techniques~\cite{wei2022emergent}.
However, LLMs struggle to achieve satisfactory performance when it comes to higher-level reasoning~\cite{stechly2023gpt,dziri2023faith} and planning~\cite{valmeekam2023can,bubeck2023sparks} tasks.
Second, most LLM-based agents avoid fine-tuning LLMs on task-specific data to preserve the model's generality and prevent over-fitting.
This strategy complicates the utilization of existing task-specific data and expertise, as well as the alignment of task scenarios with input-output formats, data distribution, and human preferences.  

To address the limitations of LLMs in complex reasoning, we distinctly separate reasoning tasks into two systems based on the dual-process theory~\cite{wason1974dual} and propose an external \Thinker module to enhance the reasoning capabilities of LLM-based agents.
In our framework, LLMs are responsible for \emph{System-1} reasoning tasks involving intuitive thinking, such as basic NLP interactions, common-sense, and symbolic reasoning,
while the \Thinker handles \emph{System-2} reasoning that requires complex logical analysis, deep understanding of domain-specific knowledge, and strategic planning in specialized tasks.
We establish a communication protocol between LLMs and the \Thinker through language-based features and instructions.
Unlike augmenting LLMs with cumbersome prompt engineering, the \Thinker is directly optimized with knowledge from databases and trained using supervised and reinforcement learning techniques, thus enhancing the LLM-agent's performance and domain alignment without compromising LLM's generality.

We select the 9-player Werewolf game as a proving ground for the proposed framework.
Werewolf is a popular social deduction game, current AI systems fall short when compared to even moderate human players in this domain.
% that requires dual-system reasoning and advanced NLP capabilities, highlighting the necessity for integrating LLMs with external reasoning modules.
\emph{System-1} reasoning tasks in Werewolf encompass natural language understanding and generation of players' statements, as well as the adept use of game-specific jargon.
Meanwhile, the hidden roles necessitate complex strategic thinking such as identity concealment, and sophisticated communication involving deception and disguise, which fall under \emph{System-2} reasoning.
This duality creates a significant gap between the players' actual statements and their true intentions,
making Werewolf an ideal testbed for assessing the advanced reasoning capabilities of LLM agents.

We have collected \num{18800} real human game sessions and analysed the primary patterns behind human speeches.
Informed by these patterns, we design language-based features for speech understanding and instructions for speech generation.
The \Thinker module is optimized by imitation learning, reinforcement learning (RL) from fictitious self-play~\cite{heinrich2015fictitious}, and population-based training~\cite{jaderberg2017population}, to output reasonable game actions and LLM speech instructions.
We compare our approach with GPT3.5/4 methods using Least-to-Most (LtM) prompting~\cite{zhou2022least} from three aspects: deductive reasoning and decision-making, human evaluation of speech generation, and online evaluation of a complete game.
Experiments demonstrate that the integration of an external \Thinker module substantially enhances the reasoning and generation capability of LLMs.
Further, we fine-tune a smaller LLM model (6B)~\cite{du2021glm} to better align real human speech styles and preferences,
outperforming GPT4 in the majority of evaluative scenarios.
Our primary contributions include:
\begin{itemize}[leftmargin=*]
\setlength{\itemsep}{0pt}
\item We propose an external \Thinker module to enhance the reasoning capabilities of LLM agents and demonstrate it by a Werewolf AI that surpasses GPT4 in real gameplay.
\item We release a dataset\footnote{\url{https://github.com/boluoweifenda/werewolf}} of \num{18800} Werewolf game sessions, which represents the largest known dataset for social deduction games to date.
\end{itemize}

\begin{figure*}[t]
\begin{center}
\includegraphics[width=\textwidth]{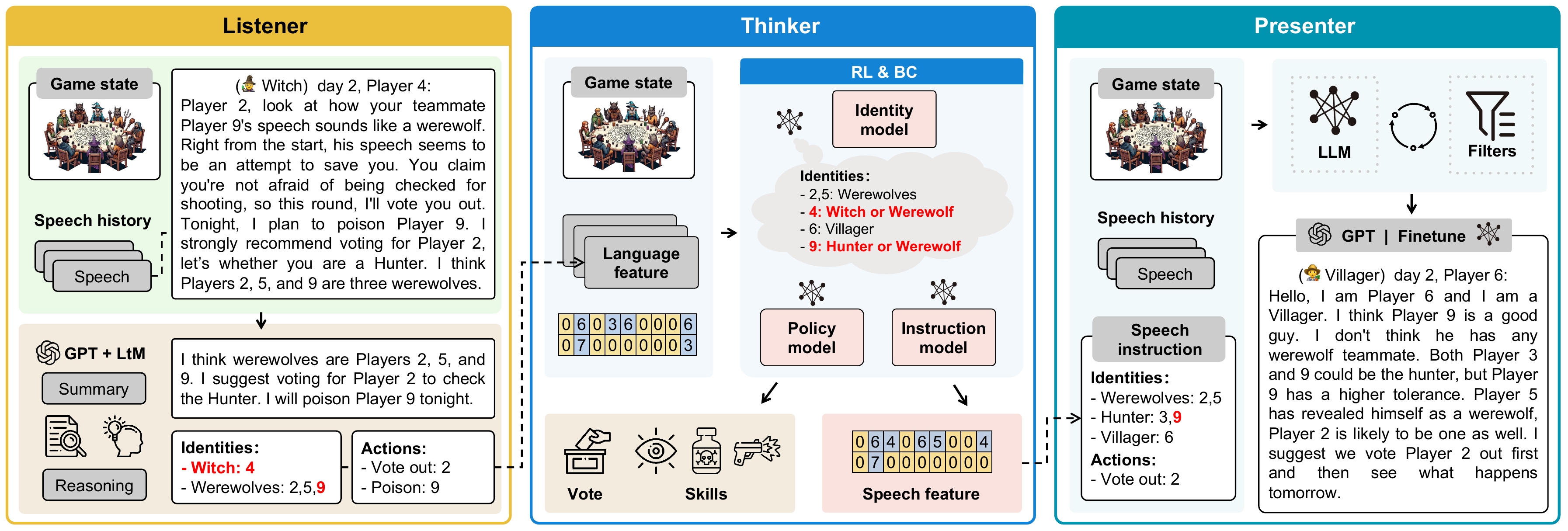} % rebuttle 修订
\end{center}
\vspace{-5pt}
\caption{Overall processing framework and modules in the Werewolf implementation.} 
\label{fig:werewolf_framework}
\end{figure*}

\section{Related Work}
\label{sec:related_work}

\textbf{Enhance Reasoning in LLMs.} 
\label{sec:enhance_reasoning_and_planning}
Several approaches bypass the need for intricate prompt engineering mentioned in the Introduction.
For instance, LLM+P~\cite{liu2023llm} employs an external planner to address long-horizon robot planning challenges.
A different approach~\cite{zhang2023building} heuristically designs a low-level planner to manage primitive control actions.
RAG~\cite{lewis2020retrieval} combines pre-trained parametric-memory generation models with a non-parametric memory to improve performance on knowledge-intensive tasks. 
Regarding the fine-tuning of LLMs,
Galactica~\cite{taylor2022galactica} is trained on a scientific dataset that emphasizes detailed reasoning processes.
WebGPT~\cite{nakano2021webgpt} utilizes human feedback to fine-tune GPT-3, enabling it to answer long-form questions within a textual web-browsing context.
Toolformer~\cite{schick2023toolformer} fine-tunes LLMs to use external tools in a self-supervised manner with human demonstrations.
OpenAGI~\cite{ge2023openagi} implements RL from feedback in open-ended tasks to refine the LLM's planning strategy.
Cicero~\cite{meta2022human} fine-tunes LLMs to generate dialogue controlled by a strategic reasoning module in the game Diplomacy.
However, the control signals in Cicero (planned actions) are insufficient to convey the complex language dynamics (both understanding and generation) in the Werewolf game.

\textbf{AI for Social Deduction Games.}
DeepRole~\cite{serrino2019finding} combines counterfactual regret minimization (CFR) with deep value networks in the non-speech five-player Avalon game.
Hidden Agenda~\cite{kopparapu2022hidden} presents a two-team, non-speech social deduction game in a 2D environment.
A system comprising three LLM-powered interfaces is created~\cite{zhu2023calypso} to aid gameplay in Dungeon Master.
Regarding AI in werewolf games, bootstrap aggregating and weighted ensemble learning have been employed to improve voting strategies~\cite{khan2022novel}.
\cite{brandizzi2021rlupus} proposes an RL framework to analyze the influence of diverse communication behaviors among agents.
One Night Ultimate Werewolf~\cite{eger2019study} explores human responses to various deliberation strategies.
In the five-player werewolf game, \cite{wang2018application} builds a deep-Q network to decide whom to trust or kill.
Deep Wolf~\cite{shibata2023playing} fine-tunes a RoBERTa-like pretrained model with 48 game logs to construct a value network given the current game state, human speeches, and candidate actions. 
The seven-player version is explored with RL and LLMs in~\cite{xu2023language,xu2023exploring}.
Our approach differs from previous studies in two fundamental ways:
First, we employ the \Thinker to execute complex \emph{System-2} reasoning,
in contrast to the reasoning approach of LLM in~\cite{xu2023language}, which generates candidate results for the RL model to select and mitigate biases.
Second, by collecting and leveraging authentic game sessions and speech data, we aim for closer alignment with real-world scenarios and human interaction patterns.

\section{Methods}
\label{sec:methods}
We introduce an innovative framework that synergizes LLMs with an external module for reasoning and decision-making, referred to as the \Thinker, devised to augment LLM-based agents with sophisticated reasoning abilities.
To bridge the communication between \Thinker and LLMs, we introduce a protocol through structured features and prompt instructions.
The framework is thus decomposed into three processing components:

\begin{itemize}[leftmargin=*]
\setlength{\itemsep}{0pt}

\item The \textbf{\Listener} serves as the primary interface for natural language understanding.
It processes language inputs, engages in intuitive \emph{System-1} reasoning, and transforms the information into structured language features that the \Thinker can interpret.

\item The \textbf{\Thinker} functions as the cognitive core of the framework.
Utilizing language features provided by the \Listener, it specializes in \emph{System-2} reasoning tasks that require deep logical analysis and domain-specific knowledge.
The \Thinker produces policies such as planning and actions, and generates strategic instructions for the \Presenter.

\item The \textbf{\Presenter} functions as the system's articulator.
It generates coherent and contextualized language output that aligns with the current environment state, guided by the strategic instructions from the \Thinker.
The \Presenter ensures that the generated language is logical, rational, consistent, and free from hallucinations.

\end{itemize}

To demonstrate the effectiveness of our framework, we apply it to the complex social deduction game Werewolf.
The remainder of this section will detail the implementation within the game environment, which necessitates deductive reasoning, speech understanding and generation, as illustrated in Figure~\ref{fig:werewolf_framework}.

\subsection{Data preparation}
\label{sec:data_preparation}
We collected data from the 9-player standard mode Werewolf game hosted on the Fanlang platform\footnote{\url{https://www.wolfkills.com/}}.
The specific rules of the game are detailed in Appendix~\ref{sec:gamerule}.
We recorded real-time video in spectator mode for approximately \num{18800} game sessions, which equates to around \num{7000} hours of gameplay and \num{6000} hours of speech.
Furthermore, we enriched our dataset with a Werewolf domain-specific corpus comprising nearly $1.4$ million characters sourced from web-crawled game strategies and OCR-processed Werewolf literature.
Each recorded session includes both the game state data and the audio of players' speeches.
We captured exhaustive game state details, such as historical skills and voting results, by utilizing an automated testing framework\footnote{\url{https://github.com/appium/appium}}.

We deployed the Paraformer~\cite{gao22b_interspeech} model for Automatic Speech Recognition (ASR) of human speech audio.
To improve recognition accuracy, especially for frequently used Werewolf-specific terms, we crafted a hot word list from the domain corpus and utilized context biasing methods~\cite{zhao2019shallow}.
Furthermore, we annotated approximately \num{127} hours of Werewolf speech data and performed supervised fine-tuning on the Paraformer model.
The character error rate of ASR for Werewolf speeches was reduced from $4.5\%$ to $3.7\%$. 
We refer to the dataset hereafter as \fanlang, and a thorough analysis of the dataset is in Appendix~\ref{sec:fanlang9}.

\subsection{Listener}
\label{sec:listener}
\begin{figure*}[t]
\begin{center}
\includegraphics[width=1.0\textwidth]{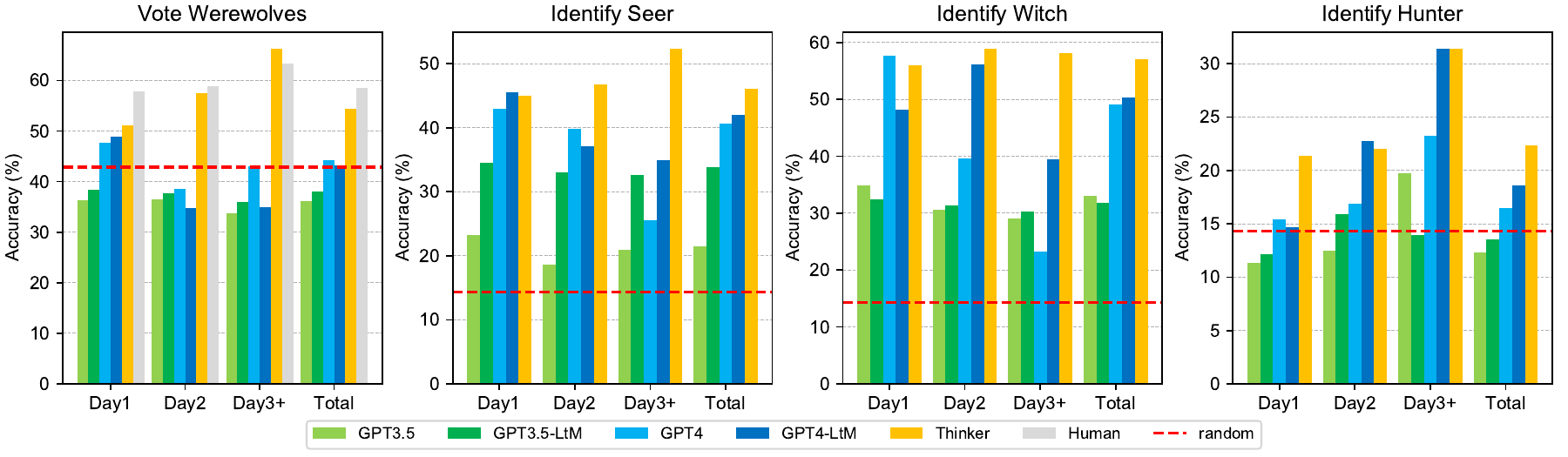}
\end{center}
\vspace{-10pt}
\caption{Voting and identification accuracy evaluating the reasoning capability from the perspective of villagers.
The random baseline is calculated as total\_role\_number/total\_hidden\_players, i.e., 3/8 or 1/8} 
\label{fig:nlu}
\end{figure*}

In the game of Werewolf, the complexity of speeches arises from players concealing their identities. 
Werewolves make deceptive statements to disguise themselves as the "Good" faction.
Conversely, the "Good" faction strives to discern werewolves by deducting from historical speeches and actions while providing rational and credible statements. 
This dynamic creates a significant gap between the players' actual statements and their true intentions (see Figure~\ref{fig:thinker_induced_example}).
The \Listener aims to capture relevant insights from actual statements without speculating on their hidden motives or truthfulness.
To tackle this, we introduce dual-phase processing:

\textbf{Synthesize and summarize}: Human players' speeches on the Fanlang platform are characterized by an information overload that includes a tangled mix of context, lengthy and redundant content, and colloquial ramblings, alongside complex logic that encompasses quotations, rhetorical questions, hypotheses, and empathetic thinking, creating a rich and intricate web of discourse.
Moreover, the accumulation of historical speeches often exceeds \num{10}K tokens (see Figure~\ref{fig:speech_tokens_distribution}), making it difficult for LLMs to directly infer information and process deductive reasoning (see Figure~\ref{fig:nlu}).
Inspired by the Least-to-Most (LtM) prompting~\cite{zhou2022least}, 
we first prompt the LLM to generate a textual summary of fewer than \num{200} words for each single statement, retaining only critical information that the speaker intends to express.

\textbf{Reasoning and feature extraction}: The LLM discerns and delineates key information from these summaries and generates
a JSON-style reasoning result of the speech, containing pairs of player IDs with their attributes.
Then the result is tokenized and categorized into specific language features, as detailed in Appendix~\ref{sec:fanlang9}.
For an $N$-player werewolf game, we define $M$ different attributes, which encompass various aspects of a player, e.g., identities, actions, and historical skills.
From the historical collection of all speeches $\mathcal{H}$, a player's single speech $\mathbf{S}$ may include descriptions of all the players in the game, the language feature can be presented by a matrix $\mathbf{F} \in \mathbb{Z}^{N \times M}$:
\begin{equation}
\label{eqn_feature}
    \mathbf{F}=[\mathbf{f}_{1},\mathbf{f}_{2},...,\mathbf{f}_{N}]^T,
\end{equation}
where $\mathbf{f_n}=[f_{n1}, f_{n2},..., f_{nM}]^T, n=1,2,...,N$ and $f_{nm}\in \mathcal{V}_m, \forall n=1,2,...,N$ and $m=1,2,...,M$.
Here $\mathcal{V}_m$ signifies the set of the potential values that the $m$-th attribute can assume.

An example of summary and language feature is illustrated in Figure~\ref{fig:werewolf_framework}.
Ablation studies in Appendix~\ref{sec_ablation} indicate that solely predicting future actions, as done in Cicero~\cite{meta2022human}, will omit crucial identity accusations, leading to substantial information loss that is detrimental in the Werewolf game.
Aside from directly prompting GPT3.5 and GPT4 to generate language features, we also extract \num{260}K speech instances from the \fanlang dataset, label the speech-feature pairs with GPT3.5, and finetune the ChatGLM-6B~\cite{du2021glm} model to perform the same reasoning task for practical efficiency.
To ensure the output format of language features, we also include a post-processing filter for GPTs and the fine-tuned model.
The detailed prompts for summary, reasoning, fine-tuning, and post-filtering are provided in Appendix~\ref{sec:llm_prompting_for_nlu}.

\subsection{Thinker}
\label{sec:thinker}
The primary objective of the \Thinker module is to tackle complex \emph{System-2} reasoning tasks.
In the game of werewolf, it analyze the underlying intentions and strategic implications behind players' public speeches.
In contrast to LLMs, which typically depend on complex prompt engineering for scenario adaptation, the \Thinker module distinguishes itself by its capacity to directly harness knowledge from databases and various optimizing techniques.
This ability allows the \Thinker to internalize human-like decision-making patterns and strategic speech intricacies that are crucial for navigating the complex dynamics in this game.

The speech instruction $\mathbf{I} \in \mathbb{Z}^{N \times M}$
follows the same structure as the language feature outlined in Equation~\ref{eqn_feature}, which can be viewed as a multi-label classification problem
and decomposed into multiple single-class classifications for each attribute $f_{nm}$.
Therefore, the generation of a speech instruction is converted into {$N \times M$} actions, aligning the same training algorithm as for game actions.
The optimization of the \Thinker module comprises two phases: imitation learning and RL.
For imitation learning, we utilize human data and the Behavioral Cloning (BC)~\cite{torabi2018behavioral} loss as:
\begin{equation}
\label{eqn_bc}
\mathcal{L}_{\text{BC}}(\theta) = -\mathbb{E}_{s, a \sim \mathcal{D}} [\log \pi_{\theta}(a|s)],
\end{equation}
where $\mathcal{D}$ denotes the dataset of human action $a$ (or decomposed speech attribute), state $s$, and $\pi_{\theta}$ is the policy parameterized by $\theta$.
For the RL phase, we employ Proximal Policy Optimization (PPO)~\cite{schulman2017proximal} and a distributional training framework~\cite{ye2020towards}:
\begin{equation}
\label{eqn_ppo}
\mathcal{L}_{\text{RL}}(\theta) = - \mathbb{E}_{s, a \sim \pi_{\theta'}}
\left[\frac{\pi_{\theta}(a|s)}{\pi_{\theta'}(a|s)}A^{\pi_{\theta}}(s,a) \right],
\end{equation}
where $\theta'$ is the parameters of an old policy, and $A^{\pi_{\theta}}(s,a)$ is the advantage with respect to policy $\pi_{\theta}$.

In addition, we incorporate an auxiliary task that predicts all players' identities.
This task serves to reveal the model's true deduction, which may contrast with the generated speech instructions.
We denote the cross-entropy loss function as $\mathcal{L}_{\text{id}}(\phi)$ with parameter $\phi$, which is labeled by human data or RL environment in a self-supervised manner.
The overall training objective of the \Thinker is formulated as:
\begin{equation}
\label{eqn_all}
\mathcal{L} = \alpha \mathcal{L}_{\text{BC}}(\theta) + \mathcal{L}_{\text{RL}}(\theta) + \beta \mathcal{L}_{\text{id}}(\phi),
\end{equation}
where $\alpha$ and $\beta$ are weighting coefficients.

Given the adversarial nature of the game, it is crucial to maintain a balanced win rate between the two opposing factions during training.
To this end, we deploy distinct models for the werewolf and the "Good" factions. 
We find that optimizing werewolves' speech instruction is much more challenging, as they need to mimic the "Good" faction's speech and master the art of disguise and deception.
To mitigate this, we draw inspiration from Generative Adversarial Networks~\cite{NIPS2014_5ca3e9b1} and adjust the training iterations, $n_{\text{werewolf}}:n_{\text{goods}} = 5:1$.
To prevent actions and speech strategies from converging to a single pattern, we employ population-based training~\cite{jaderberg2017population} with a population size of \num{4}.
We also introduce fictitious self-play~\cite{heinrich2015fictitious}, where in each game an average of $3$ players employ the latest models, while the remaining $6$ players use models randomly selected from the most recent $500$ checkpoints.
Further details on hyperparameters, reward shaping, and model structures are in Appendix~\ref{sec_implementation}.

\subsection{Presenter} 
\label{sec:presenter}
The generation of players' public speeches is a pivotal component in the Werewolf game, which significantly impacts the game's outcome due to its strategic importance and influence on other players' actions.
The quality of speech generation encompasses several critical aspects:
(1) The strategy articulated within the speech should align with the player's role and the current state of the game.
(2) Speeches need to adhere to the logical framework of the game, correlating with historical speeches and actions, making them sound and convincing.
(3) Speeches must fit the stylistic environment of the Werewolf game.
Detailed evaluation metrics can be found in Appendix~\ref{eval_nlg}.

The \Thinker handles only the first aspect of speeches,
providing a foundational stem for the \Presenter, such as the Witch's decision to report the previous night's rescue, as shown in Figure~\ref{fig:werewolf_framework}. 
Then the \Presenter crafts a complete speech that incorporates necessary contextual information relevant to the game state and historical speeches.
The \Presenter module has two fundamental objectives:
\begin{itemize}[leftmargin=*]
\setlength{\itemsep}{0pt}
\item \textbf{Controllability}: It must align with the strategic instructions provided by the \Thinker.
\item \textbf{Quality}: The generated speech should be logical, persuasive, and aligned with human players' preferences.
\end{itemize}
To achieve these objectives, the \Presenter leverages the capabilities of LLMs by incorporating the \Thinker's strategic instructions and the game state directly into the prompt, enabling LLMs to generate \Thinker-induced speeches.
The template for the prompt is provided in Appendix~\ref{sec:llm_prompt_for_nlg}.
Additionally, as with the \Listener module, we fine-tune the ChatGLM-
6B as a domain-specific Werewolf speech LLM.
We inverse the 260K speech-feature pairs in the finetuning of \Listener: the language feature $\mathbf{F}$ generated by the \Listener now serves as the hindsight speech instruction $\mathbf{I}$, and the actual speech $\mathbf{S}$ serves as output labels.

We observed that LLMs often fail to follow prompts, and even fine-tuned models exhibit hallucinations and inaccuracies.
Taking inspiration from the Cicero~\cite{meta2022human} approach, we introduce additional filtering steps. 
We use the \Listener module to perform further reasoning on the generated speeches to produce language features, which we then compare for similarity to original speech instructions.
For expressions of the speaker's own attributes, the filter requires an exact match. 
For expressions pertaining to the attributes of others, the content indicated in the speech instructions must be consistent.
For parts not mentioned in the instructions, the filter allows the \Presenter some leeway in cases of hallucinations.
The speech generation process iterates until it successfully meets the filter criteria or exceeds the maximum number of allowed attempts.
When the maximum is reached without successful compliance, a speech is generated based on rules that take into account the player's roles, historical skills, and identity predictions.

\section{Experiments}
\label{sec:experiments}
We assess the performance of our method by comparing it against several baselines and ablative variants.
The models involved in the following experiments include:
\begin{itemize}[leftmargin=*]
\setlength{\itemsep}{0pt}
\item \textbf{GPT3.5/4}: GPT3.5 and GPT4 are directly applied to generate end-to-end action decisions and speeches.
For GPT3.5, we use the model named \emph{gpt-35-turbo-16k} and version \emph{0613}.
For GPT4, we apply model name \emph{gpt-4} and version \emph{1106-Preview}.
We prompt GPTs with basic game rules, explanations of typical game jargon, and comprehensive game information, including visible game states, legal actions, and speech text converted by ASR.
Examples of the prompts can be found in Appendix~\ref{sec:llm_prompting_for_nlu}.

\item \textbf{GPT3.5/4-LtM}: We allow GPTs to first summarize each speech, as described in Section~\ref{sec:listener}, and then generate actions and speeches according to the game information and speech summaries.

\item \textbf{GPT3.5/4-T}: GPTs serve as the \Listener and \Presenter modules, and our proposed \Thinker module is integrated for generating actions and speech instructions.

\item \textbf{Finetune-T}: We replace GPTs with a 6B LLM fine-tuned on the \fanlang dataset in both the \textit{Listener} and \Presenter modules, while the \Thinker remains the same as in GPT3.5/4-T.
It is for practical efficiency concern, our framework does not necessitate the finetuning of LLMs.

\end{itemize}

\subsection{Deductive Reasoning}
\label{exp_nlu}
We evaluate the reasoning capabilities of various models, which encompass understanding and comprehending of both the game state and the historical speeches, i.e., how the models think of the current game status.
We extract \num{300} games from the \fanlang dataset as the test set. 
Models are required to identify special roles (Seer, Witch, and Hunter) and vote for the most likely werewolf, from the perspective of villagers at the first round of voting each day.
Given that villagers have limited access to information and must engage extensively in deductive reasoning within the game, this task represents a stringent test of the models' reasoning capabilities.
The test set encompasses approximately \num{1200} evaluation instances.
For the \Thinker, we utilize its action decision as the result for werewolf voting, and identities predicted by the auxiliary task as results for special roles.
We assume that human players in the test set who are villagers would vote for the most likely werewolf.
Therefore, their voting choices are listed as a reference but their judgments about other players' identities remain unknown.

The accuracy results are shown in Figure~\ref{fig:nlu}.
In terms of voting werewolves, human players have the highest accuracy and the \Thinker is closest to human players.
The \Thinker module closely mirrors human performance, notably outperforming direct reasoning methods using GPTs in the identification of werewolves and other special roles.
LtM prompting improves GPT3.5's performance, particularly in the identification of the Seer, suggesting benefits in processing complex and lengthy speech contexts.
However, the marginal gains for GPT4-LtM over GPT4 indicate that GPT4's inherent improvements in handling extensive texts make it less reliant on speech summaries and more dependent on game state experiences.
We observe that in human gameplay, Seers and Witches often disclose their roles, aiding GPTs in outperforming random baselines, while Hunters and werewolves typically conceal their roles, resulting in GPTs' performance aligning with random guessing.
Notably, the accuracy of GPTs tends to decline over successive days except for the Hunter, whereas the \Thinker's accuracy improves. 
This pattern suggests that although GPTs initially benefit from role disclosures on the first day, they may be hindered by the extensive speeches in subsequent days.

\begin{figure}[htbp]
\begin{center}
\includegraphics[width=0.9\linewidth]{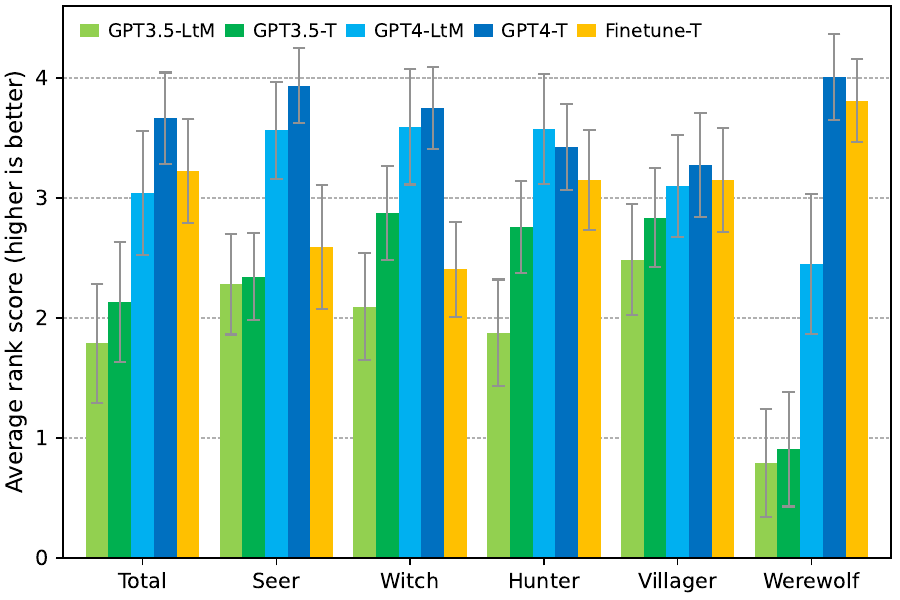}
\end{center}
\vspace{-8pt}
\caption{Human preference score for generated speeches grouped by identities.} 
\label{fig:nlg}
\end{figure}
\vspace{-4pt}

\begin{figure*}[t]
\begin{center}
\includegraphics[width=.85\textwidth]{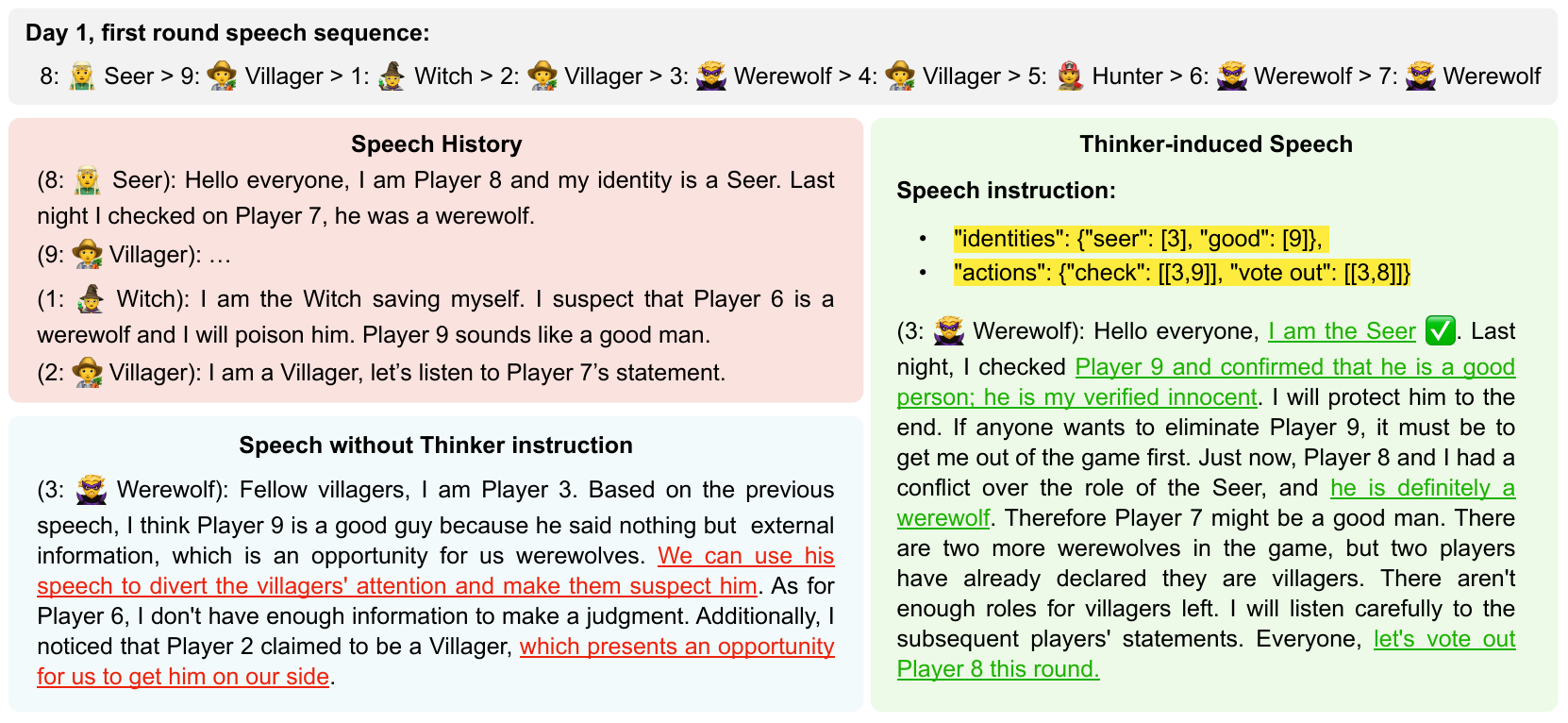}
\end{center}
\vspace{-10pt}
\caption{An example comparison of speeches with and without strategic instruction.} 
\label{fig:thinker_induced_example}
\end{figure*}

\subsection{Thinker-induced Speech generation}
\label{sec_exp_nlg}
We investigate the speech generation capabilities of various models. 
Using the same \num{300} complete games from Section~\ref{exp_nlu}, we extracted around \num{400} speech sessions that span a diverse range of roles, times of day, and speech types (first/second round speech, last words).

Models are assigned the task of generating speeches based on the current game state and all players' historical speeches, with detailed prompts for GPTs available in Appendix~\ref{sec:llm_prompt_for_nlg}.
Given the effectiveness of LtM prompting, we excluded GPTs without LtM prompting in subsequent experiments.
For GPTs-T and Finetune-T settings, speech instructions are derived from the \Thinker and incorporated into the prompts.
To evaluate models' single-shot generation ability, we do not adopt post filtering process for generated speeches in this evaluation, which yielded approximately \num{2000} speeches for five models.
To evaluate the quality of generated speeches, we recruited 10 human evaluators who are well-familiar with the Werewolf game. 
For each session, generated speeches are presented in a randomized order to ensure that evaluators are unaware of the model behind each speech.
Evaluators are required to rank the speeches and detect obvious legal errors according to the criteria detailed in Appendix~\ref{eval_nlg}.

The evaluation results are shown in Figure~\ref{fig:nlg}.
Considering total scores, GPTs induced by the \Thinker outperform their LtM-prompting counterparts, demonstrating that speech instructions significantly enhance speech quality.
Moreover, the fine-tuned 6B model surpasses GPT4 with prompting methods in speech generation capability.
Regarding scores for specific roles, the \Thinker's contribution over GPT3.5 is somewhat limited for the Seer, whose speeches are relatively straightforward, needing only to report inspections from the previous night.
The assessment of villagers' speeches is inherently complex due to their limited available information, which is reflected in the minimal rank score differences observed among the models for this role.
In contrast, rank score differences are most obvious for werewolves.
This disparity stems mainly from the low legality of werewolf speeches, which often inadvertently reveal their identity—a critical error as outlined in Appendix Table~\ref{tab_nlg}.
Notably, GPT3.5 struggles to adhere to instructions that advise against self-incrimination, while GPT4 demonstrates a more sophisticated ability to disguise itself, especially when induced by the \Thinker's strategic instructions.
An example speech is presented in Figure~\ref{fig:thinker_induced_example}.

\begin{table*}[]
\centering
\caption{Online evaluation results showcasing the performance of 9 AIs using 5 different models and 3 combinations.
Results are presented in the format: win rate | Behavior Score.}
\begin{tabular}{
p{1.85cm}
p{2.0cm}<{\centering}
p{2.0cm}<{\centering}
p{2.0cm}<{\centering}
p{2.0cm}<{\centering}
p{2.0cm}<{\centering}
p{2.0cm}<{\centering}
% p{2.1cm}<{\centering}
}
\toprule
Method & Total & Seer & Witch & Hunter & Villager & Werewolf \\
\midrule
GPT3.5-LtM & 36.7\% | $-$0.21 & 25.6\% | $+$0.16 & 23.1\% | $-$0.51 & 29.9\% | $-$0.21 & 30.8\% | $-$0.42 & 53.4\% | 0.00 \\
GPT3.5-T & 47.4\% | $-$\textbf{0.05} & 38.3\% | $+$0.27 & \textbf{41.0}\% | $-$\textbf{0.14} & 36.4\% | $-$\textbf{0.12} & 33.8\% | $-$0.18 & 68.6\% | 0.00  \\
Finetune-T & \textbf{50.3}\% | $-$0.06  & \textbf{38.8}\% | $+$\textbf{0.33} & 39.8\% | $-$0.18 & \textbf{37.0}\% | $-$0.29 & \textbf{39.1}\% | $-$\textbf{0.11} & \textbf{74.4}\% | 0.00 \\
\midrule
GPT4-LtM & 37.9\% | $-$\textbf{0.01} & 21.9\% | $+$0.25 & 18.6\% | $-$0.25 & 19.4\% | $-$\textbf{0.06} & 20.3\% | $-$\textbf{0.00} & 73.6\% | 0.00 \\
GPT4-T & 41.1\% | $-$0.02  & 20.4\% | $+$0.25 & 23.2\% | $-$\textbf{0.10} & \textbf{23.9}\% | $-$0.09 & 22.5\% | $-$0.09 & 78.4\% | 0.00 \\
Finetune-T & \textbf{43.1}\% | $-$0.04  & \textbf{24.2}\% | $+$\textbf{0.27} & \textbf{24.6}\% | $-$0.15 & 23.4\% | $-$0.15 & \textbf{23.9}\% | $-$0.11 & \textbf{81.4}\% | 0.00 \\
\midrule
GPT3.5-LtM & 33.0\% | $-$0.22 & 14.4\% | $+$0.12 & 20.4\% | $-$0.46 & 20.7\% | $-$0.57 & 21.6\% | $-$0.33 & 57.0\% | 0.00 \\
GPT3.5-T & 45.0\% | $-$0.07 & \textbf{33.6}\% | $+$\textbf{0.29} & 32.2\% | $-$0.13 & 30.4\% | $-$0.17 & 27.6\% | $-$0.20 & 75.8\% | 0.00  \\
GPT4-LtM & 42.5\% | $-$\textbf{0.03} & 29.8\% | $+$0.27 & 22.2\% | $-$0.18 & 27.0\% | $-$0.20 & 28.7\% | $-$\textbf{0.04} & 71.9\% | 0.00 \\
GPT4-T & \textbf{46.3}\% | $-$0.05  & 28.6\% | $+$0.28 & \textbf{34.5}\% | $-$\textbf{0.11} & \textbf{31.5}\% | $-$\textbf{0.08} & 28.0\% | $-$0.18 & \textbf{79.9}\% | 0.00 \\
Finetune-T & 45.9\% | $-$0.06  & 29.1\% | $+$0.25 & 28.3\% | $-$0.16 & 29.2\% | $-$0.21 & \textbf{32.4}\% | $-$0.14 & 78.0\% | 0.00 \\
% Human & &&&&& \\
\bottomrule
\end{tabular}
\label{tab_eval}
\end{table*}

\subsection{Online Evaluation} 
\label{sec:online_evaluation} 
Lastly, we conduct online evaluations to assess the overall performance in a real-world gameplay setting, which involves the five models in the speech generation evaluation: GPT3.5/4-LtM, GPT3.5/4-T and Finetune-T.
As Werewolf is a multiplayer imperfect-information game, the skill level of participants can significantly affect the evaluation results.
Therefore, we devise three model combinations, within which models are randomly and repeatedly selected to simulate a nine-player game.
We conducted approximately \num{600} rounds for each combination to ensure robust testing results.
Given the inherent randomness of outcomes in Werewolf, we also calculate the Behavior Score, a typical metric used in Werewolf competitions\footnote{\url{https://langrensha.163.com/20230313/31014\_1077578.html}}.
A comprehensive breakdown of the Behavior Score is provided in Table~\ref{tab:behavior_score}.

Table~\ref{tab_eval} summarizes the results, revealing that the integration of the \Thinker module markedly boosts the win rates of both GPT3.5 and GPT4 in all three combinations. 
The performance of the Finetune-T model closely aligns with that of GPT4-T. 
In terms of Behavior Score, the \Thinker contributes substantial improvements across all roles for GPT3.5. 
For GPT4, notable benefits are observed particularly for the Witch and Hunter roles.
The Behavior Score metric assigns significant weight to the witch's poisoning and the hunter's shooting decisions, which correlates with the \Thinker's ability to enhance werewolf detection and subsequently improve these scores.
Another finding is that the combination of GPT4s and Finetune-T models results in the highest win rate for the werewolves.
This outcome primarily stems from the conservative nature of GPT4-LtM in role identification, which makes it more cautious in voting and skill usage as the "Good" faction. 

Furthermore, we incorporate \num{13} human players to evaluate AI performance against human strategy.
We find that the issue of werewolf identity exposure, as mentioned in Section~\ref{sec_exp_nlg}, significantly impedes the game experience of human players.
As a result, participants play alongside four instances each of GPT4-T and Finetune-T models across \num{200} game rounds, and the post-filtering process for generated speeches is adopted in this setting.
As can be seen in Table~\ref{tab_human}, human players exhibit no significant win rate advantage, suggesting that the AI's speeches and actions do not exhibit exploitable weaknesses.
Moreover, when compared with the results in Table~\ref{tab_eval}, we note a relative decrease in the werewolves' win rate in games involving human players, highlighting the ongoing challenges related to identity concealment.
Although AI-managed werewolves might convincingly deceive other AI players, human players often find them suspicious.
A typical example is that werewolves tend to act in groups, such as unanimously voting for \emph{Player\_3}.

\begin{table}[h]
\centering
\caption{Online evaluation win rates with 1 human and 8AIs.}
\begin{tabular}{
p{1.6cm}
p{1.5cm}<{\centering}
p{1.5cm}<{\centering}
p{1.5cm}<{\centering}
% p{2.1cm}<{\centering}
}
\toprule
Method & Total & Goods & Werewolves \\
\midrule
GPT4-T     & 46.9\% & 37.3\% & 65.0\% \\
Finetune-T & 45.3\% & 36.0\% & 62.6\% \\
Human      & 40.5\% & 35.3\% & 59.4\% \\
\bottomrule
\end{tabular}
\label{tab_human}
\end{table}

\section{Discussion and Future Work}
\label{sec:discussion}

\textbf{Transfer to other tasks:}
We use language features and speech instructions in our framework to integrate LLMs and external reasoning models.
The communication format may not be directly transferable to other tasks or domains, and the effectiveness depends on the richness of these features and instructions.
Future work will aim to develop more generalized and flexible methods, e.g., implicit hidden vectors in a data-driven manner, which would offer better transferability but at the expense of interpretability and controllability.

\textbf{Evaluation of 8 humans with 1 AI:}
Our evaluations primarily involved games featuring either AI vs AI or one human player competing against multiple AIs.
Evaluating an AI in a majority-human player setting presents challenges due to the highly interactive nature of the game and the variability in human players' speech strategies and behaviors.

\textbf{Interpretability:} 
While our framework improves the reasoning capabilities of LLMs, the reasoning processes in the \Thinker module may not be easily interpretable to humans.
We explicitly introduce the identity prediction task to reveal how the \Thinker think of other players.
Future work could explore methods for further improving the interpretability and transparency of our framework.

\section{Conclusion}
\label{sec:conclusion}
In this paper, we introduced a novel framework for integrating LLMs with an external \Thinker, aiming to enhance the reasoning capabilities of LLM-based agents.
This approach is inspired by the dual-process theory and separates reasoning tasks into two systems: System-1, handled by LLMs, and System-2, handled by the \Thinker model.
We showcased our approach in the context of the Werewolf game, a complex social deduction game requiring language processing, intuitive thinking, and strategic planning.
Our results show that our framework can significantly improve the performance of LLMs and achieve better alignment with real-world scenarios and human preferences.
Additionally, we fine-tune a 6B model to surpass GPT4 when integrated with the \Thinker.
This paper also contributes the largest dataset for social deduction games to date, hoping to accelerate further advancements in this field.

\section*{Acknowledgements}
We thank Jian Yao, Weiming Liu, Qing Wang, Ye Tian, Zimeng Zhou, Yiming Gao, Liangzhou Wang, Kaiwen Zhu, Feiyu Liu, Jianing Shi, Fengming Zhu, Xiaoyu Yang for the human evaluation of speech generation and online games.
We thank Jian Yao, Jianing Shi and Guohua Tang for the thoughtful discussion.

\balance
\bibliography{arxiv}

\begin{thebibliography}{59}
\providecommand{\natexlab}[1]{#1}
\providecommand{\url}[1]{\texttt{#1}}
\expandafter\ifx\csname urlstyle\endcsname\relax
  \providecommand{\doi}[1]{doi: #1}\else
  \providecommand{\doi}{doi: \begingroup \urlstyle{rm}\Url}\fi

\bibitem[Ahn et~al.(2022)Ahn, Brohan, Brown, Chebotar, Cortes, David, Finn, Fu,
  Gopalakrishnan, Hausman, et~al.]{ahn2022can}
Ahn, M., Brohan, A., Brown, N., Chebotar, Y., Cortes, O., David, B., Finn, C.,
  Fu, C., Gopalakrishnan, K., Hausman, K., et~al.
\newblock Do as i can, not as i say: Grounding language in robotic affordances.
\newblock \emph{arXiv preprint arXiv:2204.01691}, 2022.

\bibitem[Anil et~al.(2022)Anil, Wu, Andreassen, Lewkowycz, Misra, Ramasesh,
  Slone, Gur-Ari, Dyer, and Neyshabur]{anil2022exploring}
Anil, C., Wu, Y., Andreassen, A., Lewkowycz, A., Misra, V., Ramasesh, V.,
  Slone, A., Gur-Ari, G., Dyer, E., and Neyshabur, B.
\newblock Exploring length generalization in large language models.
\newblock \emph{Advances in Neural Information Processing Systems},
  35:\penalty0 38546--38556, 2022.

\bibitem[Anil et~al.(2023)Anil, Dai, Firat, Johnson, Lepikhin, Passos, Shakeri,
  Taropa, Bailey, Chen, et~al.]{anil2023palm}
Anil, R., Dai, A.~M., Firat, O., Johnson, M., Lepikhin, D., Passos, A.,
  Shakeri, S., Taropa, E., Bailey, P., Chen, Z., et~al.
\newblock Palm 2 technical report.
\newblock \emph{arXiv preprint arXiv:2305.10403}, 2023.

\bibitem[Bakhtin et~al.(2022)Bakhtin, Brown, Dinan, Farina, Flaherty, Fried,
  Goff, Gray, Hu, et~al.]{meta2022human}
Bakhtin, A., Brown, N., Dinan, E., Farina, G., Flaherty, C., Fried, D., Goff,
  A., Gray, J., Hu, H., et~al.
\newblock Human-level play in the game of diplomacy by combining language
  models with strategic reasoning.
\newblock \emph{Science}, 378\penalty0 (6624):\penalty0 1067--1074, 2022.

\bibitem[Brandizzi et~al.(2021)Brandizzi, Grossi, and
  Iocchi]{brandizzi2021rlupus}
Brandizzi, N., Grossi, D., and Iocchi, L.
\newblock Rlupus: Cooperation through emergent communication in the werewolf
  social deduction game.
\newblock \emph{Intelligenza Artificiale}, 15\penalty0 (2):\penalty0 55--70,
  2021.

\bibitem[Bubeck et~al.(2023)Bubeck, Chandrasekaran, Eldan, Gehrke, Horvitz,
  Kamar, Lee, Lee, Li, Lundberg, et~al.]{bubeck2023sparks}
Bubeck, S., Chandrasekaran, V., Eldan, R., Gehrke, J., Horvitz, E., Kamar, E.,
  Lee, P., Lee, Y.~T., Li, Y., Lundberg, S., et~al.
\newblock Sparks of artificial general intelligence: Early experiments with
  gpt-4.
\newblock \emph{arXiv preprint arXiv:2303.12712}, 2023.

\bibitem[Du et~al.(2021)Du, Qian, Liu, Ding, Qiu, Yang, and Tang]{du2021glm}
Du, Z., Qian, Y., Liu, X., Ding, M., Qiu, J., Yang, Z., and Tang, J.
\newblock Glm: General language model pretraining with autoregressive blank
  infilling.
\newblock \emph{arXiv preprint arXiv:2103.10360}, 2021.

\bibitem[Dziri et~al.(2023)Dziri, Lu, Sclar, Li, Jian, Lin, West, Bhagavatula,
  Bras, Hwang, et~al.]{dziri2023faith}
Dziri, N., Lu, X., Sclar, M., Li, X.~L., Jian, L., Lin, B.~Y., West, P.,
  Bhagavatula, C., Bras, R.~L., Hwang, J.~D., et~al.
\newblock Faith and fate: Limits of transformers on compositionality.
\newblock \emph{arXiv preprint arXiv:2305.18654}, 2023.

\bibitem[Eger \& Martens(2019)Eger and Martens]{eger2019study}
Eger, M. and Martens, C.
\newblock A study of ai agent commitment in one night ultimate werewolf with
  human players.
\newblock In \emph{Proceedings of the AAAI Conference on Artificial
  Intelligence and Interactive Digital Entertainment}, volume~15, pp.\
  139--145, 2019.

\bibitem[Gao et~al.(2022)Gao, Zhang, McLoughlin, and Yan]{gao22b_interspeech}
Gao, Z., Zhang, S., McLoughlin, I., and Yan, Z.
\newblock {Paraformer: Fast and Accurate Parallel Transformer for
  Non-autoregressive End-to-End Speech Recognition}.
\newblock In \emph{Proc. Interspeech 2022}, pp.\  2063--2067, 2022.
\newblock \doi{10.21437/Interspeech.2022-9996}.

\bibitem[Ge et~al.(2023)Ge, Hua, Ji, Tan, Xu, and Zhang]{ge2023openagi}
Ge, Y., Hua, W., Ji, J., Tan, J., Xu, S., and Zhang, Y.
\newblock Openagi: When llm meets domain experts.
\newblock \emph{arXiv preprint arXiv:2304.04370}, 2023.

\bibitem[Goodfellow et~al.(2014)Goodfellow, Pouget-Abadie, Mirza, Xu,
  Warde-Farley, Ozair, Courville, and Bengio]{NIPS2014_5ca3e9b1}
Goodfellow, I., Pouget-Abadie, J., Mirza, M., Xu, B., Warde-Farley, D., Ozair,
  S., Courville, A., and Bengio, Y.
\newblock Generative adversarial nets.
\newblock In Ghahramani, Z., Welling, M., Cortes, C., Lawrence, N., and
  Weinberger, K. (eds.), \emph{Advances in Neural Information Processing
  Systems}, volume~27. Curran Associates, Inc., 2014.

\bibitem[Heinrich et~al.(2015)Heinrich, Lanctot, and
  Silver]{heinrich2015fictitious}
Heinrich, J., Lanctot, M., and Silver, D.
\newblock Fictitious self-play in extensive-form games.
\newblock In \emph{International conference on machine learning}, pp.\
  805--813. PMLR, 2015.

\bibitem[Hu et~al.(2023)Hu, Fu, Du, Luo, Zhao, and Zhao]{hu2023chatdb}
Hu, C., Fu, J., Du, C., Luo, S., Zhao, J., and Zhao, H.
\newblock Chatdb: Augmenting llms with databases as their symbolic memory.
\newblock \emph{arXiv preprint arXiv:2306.03901}, 2023.

\bibitem[Huang \& Chang(2022)Huang and Chang]{huang2022towards}
Huang, J. and Chang, K. C.-C.
\newblock Towards reasoning in large language models: A survey.
\newblock \emph{arXiv preprint arXiv:2212.10403}, 2022.

\bibitem[Huang et~al.(2022)Huang, Abbeel, Pathak, and
  Mordatch]{huang2022language}
Huang, W., Abbeel, P., Pathak, D., and Mordatch, I.
\newblock Language models as zero-shot planners: Extracting actionable
  knowledge for embodied agents.
\newblock In \emph{International Conference on Machine Learning}, pp.\
  9118--9147. PMLR, 2022.

\bibitem[Jaderberg et~al.(2017)Jaderberg, Dalibard, Osindero, Czarnecki,
  Donahue, Razavi, Vinyals, Green, Dunning, Simonyan,
  et~al.]{jaderberg2017population}
Jaderberg, M., Dalibard, V., Osindero, S., Czarnecki, W.~M., Donahue, J.,
  Razavi, A., Vinyals, O., Green, T., Dunning, I., Simonyan, K., et~al.
\newblock Population based training of neural networks.
\newblock \emph{arXiv preprint arXiv:1711.09846}, 2017.

\bibitem[Kaplan et~al.(2020)Kaplan, McCandlish, Henighan, Brown, Chess, Child,
  Gray, Radford, Wu, and Amodei]{kaplan2020scaling}
Kaplan, J., McCandlish, S., Henighan, T., Brown, T.~B., Chess, B., Child, R.,
  Gray, S., Radford, A., Wu, J., and Amodei, D.
\newblock Scaling laws for neural language models.
\newblock \emph{arXiv preprint arXiv:2001.08361}, 2020.

\bibitem[Khan \& Aranha(2022)Khan and Aranha]{khan2022novel}
Khan, M. and Aranha, C.
\newblock A novel weighted ensemble learning based agent for the werewolf game.
\newblock \emph{arXiv preprint arXiv:2205.09813}, 2022.

\bibitem[Kopparapu et~al.(2022)Kopparapu, Du{\'e}{\~n}ez-Guzm{\'a}n, Matyas,
  Vezhnevets, Agapiou, McKee, Everett, Marecki, Leibo, and
  Graepel]{kopparapu2022hidden}
Kopparapu, K., Du{\'e}{\~n}ez-Guzm{\'a}n, E.~A., Matyas, J., Vezhnevets, A.~S.,
  Agapiou, J.~P., McKee, K.~R., Everett, R., Marecki, J., Leibo, J.~Z., and
  Graepel, T.
\newblock Hidden agenda: a social deduction game with diverse learned
  equilibria.
\newblock \emph{arXiv preprint arXiv:2201.01816}, 2022.

\bibitem[Lewis et~al.(2020)Lewis, Perez, Piktus, Petroni, Karpukhin, Goyal,
  K{\"u}ttler, Lewis, Yih, Rockt{\"a}schel, et~al.]{lewis2020retrieval}
Lewis, P., Perez, E., Piktus, A., Petroni, F., Karpukhin, V., Goyal, N.,
  K{\"u}ttler, H., Lewis, M., Yih, W.-t., Rockt{\"a}schel, T., et~al.
\newblock Retrieval-augmented generation for knowledge-intensive nlp tasks.
\newblock \emph{Advances in Neural Information Processing Systems},
  33:\penalty0 9459--9474, 2020.

\bibitem[Lin et~al.(2023)Lin, Zhao, Zhang, Wu, Ping, and
  Chen]{lin2023agentsims}
Lin, J., Zhao, H., Zhang, A., Wu, Y., Ping, H., and Chen, Q.
\newblock Agentsims: An open-source sandbox for large language model
  evaluation.
\newblock \emph{arXiv preprint arXiv:2308.04026}, 2023.

\bibitem[Liu et~al.(2023{\natexlab{a}})Liu, Jiang, Zhang, Liu, Zhang, Biswas,
  and Stone]{liu2023llm}
Liu, B., Jiang, Y., Zhang, X., Liu, Q., Zhang, S., Biswas, J., and Stone, P.
\newblock Llm+ p: Empowering large language models with optimal planning
  proficiency.
\newblock \emph{arXiv preprint arXiv:2304.11477}, 2023{\natexlab{a}}.

\bibitem[Liu et~al.(2023{\natexlab{b}})Liu, Yu, Zhang, Xu, Lei, Lai, Gu, Ding,
  Men, Yang, et~al.]{liu2023agentbench}
Liu, X., Yu, H., Zhang, H., Xu, Y., Lei, X., Lai, H., Gu, Y., Ding, H., Men,
  K., Yang, K., et~al.
\newblock Agentbench: Evaluating llms as agents.
\newblock \emph{arXiv preprint arXiv:2308.03688}, 2023{\natexlab{b}}.

\bibitem[Madaan et~al.(2023)Madaan, Tandon, Gupta, Hallinan, Gao, Wiegreffe,
  Alon, Dziri, Prabhumoye, Yang, et~al.]{madaan2023self}
Madaan, A., Tandon, N., Gupta, P., Hallinan, S., Gao, L., Wiegreffe, S., Alon,
  U., Dziri, N., Prabhumoye, S., Yang, Y., et~al.
\newblock Self-refine: Iterative refinement with self-feedback.
\newblock \emph{arXiv preprint arXiv:2303.17651}, 2023.

\bibitem[Nakano et~al.(2021)Nakano, Hilton, Balaji, Wu, Ouyang, Kim, Hesse,
  Jain, Kosaraju, Saunders, et~al.]{nakano2021webgpt}
Nakano, R., Hilton, J., Balaji, S., Wu, J., Ouyang, L., Kim, C., Hesse, C.,
  Jain, S., Kosaraju, V., Saunders, W., et~al.
\newblock Webgpt: Browser-assisted question-answering with human feedback.
\newblock \emph{arXiv preprint arXiv:2112.09332}, 2021.

\bibitem[OpenAI(2023)]{openai2023gpt}
OpenAI, R.
\newblock Gpt-4 technical report. arxiv 2303.08774.
\newblock \emph{View in Article}, 2:\penalty0 13, 2023.

\bibitem[Ouyang et~al.(2022)Ouyang, Wu, Jiang, Almeida, Wainwright, Mishkin,
  Zhang, Agarwal, Slama, Ray, et~al.]{ouyang2022training}
Ouyang, L., Wu, J., Jiang, X., Almeida, D., Wainwright, C., Mishkin, P., Zhang,
  C., Agarwal, S., Slama, K., Ray, A., et~al.
\newblock Training language models to follow instructions with human feedback.
\newblock \emph{Advances in Neural Information Processing Systems},
  35:\penalty0 27730--27744, 2022.

\bibitem[Park et~al.(2023)Park, O'Brien, Cai, Morris, Liang, and
  Bernstein]{park2023generative}
Park, J.~S., O'Brien, J., Cai, C.~J., Morris, M.~R., Liang, P., and Bernstein,
  M.~S.
\newblock Generative agents: Interactive simulacra of human behavior.
\newblock In \emph{Proceedings of the 36th Annual ACM Symposium on User
  Interface Software and Technology}, pp.\  1--22, 2023.

\bibitem[Qian et~al.(2023)Qian, Cong, Liu, Yang, Chen, Su, Dang, Li, Xu, Li,
  Liu, and Sun]{qian2023communicative}
Qian, C., Cong, X., Liu, W., Yang, C., Chen, W., Su, Y., Dang, Y., Li, J., Xu,
  J., Li, D., Liu, Z., and Sun, M.
\newblock Communicative agents for software development, 2023.

\bibitem[Qin et~al.(2023)Qin, Liang, Ye, Zhu, Yan, Lu, Lin, Cong, Tang, Qian,
  et~al.]{qin2023toolllm}
Qin, Y., Liang, S., Ye, Y., Zhu, K., Yan, L., Lu, Y., Lin, Y., Cong, X., Tang,
  X., Qian, B., et~al.
\newblock Toolllm: Facilitating large language models to master 16000+
  real-world apis.
\newblock \emph{arXiv preprint arXiv:2307.16789}, 2023.

\bibitem[Schick et~al.(2023)Schick, Dwivedi-Yu, Dess{\`\i}, Raileanu, Lomeli,
  Zettlemoyer, Cancedda, and Scialom]{schick2023toolformer}
Schick, T., Dwivedi-Yu, J., Dess{\`\i}, R., Raileanu, R., Lomeli, M.,
  Zettlemoyer, L., Cancedda, N., and Scialom, T.
\newblock Toolformer: Language models can teach themselves to use tools.
\newblock \emph{arXiv preprint arXiv:2302.04761}, 2023.

\bibitem[Schulman et~al.(2017)Schulman, Wolski, Dhariwal, Radford, and
  Klimov]{schulman2017proximal}
Schulman, J., Wolski, F., Dhariwal, P., Radford, A., and Klimov, O.
\newblock Proximal policy optimization algorithms.
\newblock \emph{arXiv preprint arXiv:1707.06347}, 2017.

\bibitem[Serrino et~al.(2019)Serrino, Kleiman-Weiner, Parkes, and
  Tenenbaum]{serrino2019finding}
Serrino, J., Kleiman-Weiner, M., Parkes, D.~C., and Tenenbaum, J.
\newblock Finding friend and foe in multi-agent games.
\newblock \emph{Advances in Neural Information Processing Systems}, 32, 2019.

\bibitem[Shibata et~al.(2023)Shibata, Miki, and Nakamura]{shibata2023playing}
Shibata, H., Miki, S., and Nakamura, Y.
\newblock Playing the werewolf game with artificial intelligence for language
  understanding.
\newblock \emph{arXiv preprint arXiv:2302.10646}, 2023.

\bibitem[Shinn et~al.(2023)Shinn, Cassano, Gopinath, Narasimhan, and
  Yao]{shinn2023reflexion}
Shinn, N., Cassano, F., Gopinath, A., Narasimhan, K.~R., and Yao, S.
\newblock Reflexion: Language agents with verbal reinforcement learning.
\newblock In \emph{Thirty-seventh Conference on Neural Information Processing
  Systems}, 2023.

\bibitem[Stechly et~al.(2023)Stechly, Marquez, and Kambhampati]{stechly2023gpt}
Stechly, K., Marquez, M., and Kambhampati, S.
\newblock Gpt-4 doesn't know it's wrong: An analysis of iterative prompting for
  reasoning problems.
\newblock \emph{arXiv preprint arXiv:2310.12397}, 2023.

\bibitem[Taylor et~al.(2022)Taylor, Kardas, Cucurull, Scialom, Hartshorn,
  Saravia, Poulton, Kerkez, and Stojnic]{taylor2022galactica}
Taylor, R., Kardas, M., Cucurull, G., Scialom, T., Hartshorn, A., Saravia, E.,
  Poulton, A., Kerkez, V., and Stojnic, R.
\newblock Galactica: A large language model for science.
\newblock \emph{arXiv preprint arXiv:2211.09085}, 2022.

\bibitem[Thoppilan et~al.(2022)Thoppilan, De~Freitas, Hall, Shazeer,
  Kulshreshtha, Cheng, Jin, Bos, Baker, Du, et~al.]{thoppilan2022lamda}
Thoppilan, R., De~Freitas, D., Hall, J., Shazeer, N., Kulshreshtha, A., Cheng,
  H.-T., Jin, A., Bos, T., Baker, L., Du, Y., et~al.
\newblock Lamda: Language models for dialog applications.
\newblock \emph{arXiv preprint arXiv:2201.08239}, 2022.

\bibitem[Torabi et~al.(2018)Torabi, Warnell, and Stone]{torabi2018behavioral}
Torabi, F., Warnell, G., and Stone, P.
\newblock Behavioral cloning from observation.
\newblock \emph{arXiv preprint arXiv:1805.01954}, 2018.

\bibitem[Valmeekam et~al.(2023)Valmeekam, Marquez, and
  Kambhampati]{valmeekam2023can}
Valmeekam, K., Marquez, M., and Kambhampati, S.
\newblock Can large language models really improve by self-critiquing their own
  plans?
\newblock \emph{arXiv preprint arXiv:2310.08118}, 2023.

\bibitem[Wang et~al.(2023)Wang, Xie, Jiang, Mandlekar, Xiao, Zhu, Fan, and
  Anandkumar]{wang2023voyager}
Wang, G., Xie, Y., Jiang, Y., Mandlekar, A., Xiao, C., Zhu, Y., Fan, L., and
  Anandkumar, A.
\newblock Voyager: An open-ended embodied agent with large language models.
\newblock \emph{arXiv preprint arXiv:2305.16291}, 2023.

\bibitem[Wang \& Kaneko(2018)Wang and Kaneko]{wang2018application}
Wang, T. and Kaneko, T.
\newblock Application of deep reinforcement learning in werewolf game agents.
\newblock In \emph{2018 Conference on Technologies and Applications of
  Artificial Intelligence (TAAI)}, pp.\  28--33. IEEE, 2018.

\bibitem[Wason \& Evans(1974)Wason and Evans]{wason1974dual}
Wason, P.~C. and Evans, J. S.~B.
\newblock Dual processes in reasoning?
\newblock \emph{Cognition}, 3\penalty0 (2):\penalty0 141--154, 1974.

\bibitem[Wei et~al.(2022{\natexlab{a}})Wei, Tay, Bommasani, Raffel, Zoph,
  Borgeaud, Yogatama, Bosma, Zhou, Metzler, et~al.]{wei2022emergent}
Wei, J., Tay, Y., Bommasani, R., Raffel, C., Zoph, B., Borgeaud, S., Yogatama,
  D., Bosma, M., Zhou, D., Metzler, D., et~al.
\newblock Emergent abilities of large language models.
\newblock \emph{arXiv preprint arXiv:2206.07682}, 2022{\natexlab{a}}.

\bibitem[Wei et~al.(2022{\natexlab{b}})Wei, Wang, Schuurmans, Bosma, Xia, Chi,
  Le, Zhou, et~al.]{wei2022chain}
Wei, J., Wang, X., Schuurmans, D., Bosma, M., Xia, F., Chi, E., Le, Q.~V.,
  Zhou, D., et~al.
\newblock Chain-of-thought prompting elicits reasoning in large language
  models.
\newblock \emph{Advances in Neural Information Processing Systems},
  35:\penalty0 24824--24837, 2022{\natexlab{b}}.

\bibitem[Wu et~al.(2022)Wu, Terry, and Cai]{wu2022ai}
Wu, T., Terry, M., and Cai, C.~J.
\newblock Ai chains: Transparent and controllable human-ai interaction by
  chaining large language model prompts.
\newblock In \emph{Proceedings of the 2022 CHI conference on human factors in
  computing systems}, pp.\  1--22, 2022.

\bibitem[Xu et~al.(2023{\natexlab{a}})Xu, Wang, Li, Luo, Wang, Liu, and
  Liu]{xu2023exploring}
Xu, Y., Wang, S., Li, P., Luo, F., Wang, X., Liu, W., and Liu, Y.
\newblock Exploring large language models for communication games: An empirical
  study on werewolf.
\newblock \emph{arXiv preprint arXiv:2309.04658}, 2023{\natexlab{a}}.

\bibitem[Xu et~al.(2023{\natexlab{b}})Xu, Yu, Fang, Wang, and
  Wu]{xu2023language}
Xu, Z., Yu, C., Fang, F., Wang, Y., and Wu, Y.
\newblock Language agents with reinforcement learning for strategic play in the
  werewolf game.
\newblock \emph{arXiv preprint arXiv:2310.18940}, 2023{\natexlab{b}}.

\bibitem[Yang et~al.(2023)Yang, Li, Wang, Lin, Azarnasab, Ahmed, Liu, Liu,
  Zeng, and Wang]{yang2023mm}
Yang, Z., Li, L., Wang, J., Lin, K., Azarnasab, E., Ahmed, F., Liu, Z., Liu,
  C., Zeng, M., and Wang, L.
\newblock Mm-react: Prompting chatgpt for multimodal reasoning and action.
\newblock \emph{arXiv preprint arXiv:2303.11381}, 2023.

\bibitem[Yao et~al.(2022)Yao, Zhao, Yu, Du, Shafran, Narasimhan, and
  Cao]{yao2022react}
Yao, S., Zhao, J., Yu, D., Du, N., Shafran, I., Narasimhan, K., and Cao, Y.
\newblock React: Synergizing reasoning and acting in language models.
\newblock \emph{arXiv preprint arXiv:2210.03629}, 2022.

\bibitem[Ye et~al.(2020)Ye, Chen, Zhang, Chen, Yuan, Liu, Chen, Liu, Qiu, Yu,
  et~al.]{ye2020towards}
Ye, D., Chen, G., Zhang, W., Chen, S., Yuan, B., Liu, B., Chen, J., Liu, Z.,
  Qiu, F., Yu, H., et~al.
\newblock Towards playing full moba games with deep reinforcement learning.
\newblock \emph{Advances in Neural Information Processing Systems},
  33:\penalty0 621--632, 2020.

\bibitem[Yin et~al.(2023)Yin, Fu, Zhao, Li, Sun, Xu, and Chen]{yin2023survey}
Yin, S., Fu, C., Zhao, S., Li, K., Sun, X., Xu, T., and Chen, E.
\newblock A survey on multimodal large language models.
\newblock \emph{arXiv preprint arXiv:2306.13549}, 2023.

\bibitem[Zhang et~al.(2023{\natexlab{a}})Zhang, Du, Shan, Zhou, Du, Tenenbaum,
  Shu, and Gan]{zhang2023building}
Zhang, H., Du, W., Shan, J., Zhou, Q., Du, Y., Tenenbaum, J.~B., Shu, T., and
  Gan, C.
\newblock Building cooperative embodied agents modularly with large language
  models.
\newblock \emph{arXiv preprint arXiv:2307.02485}, 2023{\natexlab{a}}.

\bibitem[Zhang et~al.(2023{\natexlab{b}})Zhang, Ladhak, Durmus, Liang, McKeown,
  and Hashimoto]{zhang2023benchmarking}
Zhang, T., Ladhak, F., Durmus, E., Liang, P., McKeown, K., and Hashimoto, T.~B.
\newblock Benchmarking large language models for news summarization.
\newblock \emph{arXiv preprint arXiv:2301.13848}, 2023{\natexlab{b}}.

\bibitem[Zhao et~al.(2019)Zhao, Sainath, Rybach, Rondon, Bhatia, Li, and
  Pang]{zhao2019shallow}
Zhao, D., Sainath, T.~N., Rybach, D., Rondon, P., Bhatia, D., Li, B., and Pang,
  R.
\newblock Shallow-fusion end-to-end contextual biasing.
\newblock In \emph{Interspeech}, pp.\  1418--1422, 2019.

\bibitem[Zhong et~al.(2023)Zhong, Guo, Gao, and Wang]{zhong2023memorybank}
Zhong, W., Guo, L., Gao, Q., and Wang, Y.
\newblock Memorybank: Enhancing large language models with long-term memory.
\newblock \emph{arXiv preprint arXiv:2305.10250}, 2023.

\bibitem[Zhou et~al.(2022)Zhou, Sch{\"a}rli, Hou, Wei, Scales, Wang,
  Schuurmans, Cui, Bousquet, Le, et~al.]{zhou2022least}
Zhou, D., Sch{\"a}rli, N., Hou, L., Wei, J., Scales, N., Wang, X., Schuurmans,
  D., Cui, C., Bousquet, O., Le, Q., et~al.
\newblock Least-to-most prompting enables complex reasoning in large language
  models.
\newblock \emph{arXiv preprint arXiv:2205.10625}, 2022.

\bibitem[Zhu et~al.(2023)Zhu, Martin, Head, and Callison-Burch]{zhu2023calypso}
Zhu, A., Martin, L., Head, A., and Callison-Burch, C.
\newblock Calypso: Llms as dungeon master's assistants.
\newblock In \emph{Proceedings of the AAAI Conference on Artificial
  Intelligence and Interactive Digital Entertainment}, volume~19, pp.\
  380--390, 2023.

\end{thebibliography}
\bibliographystyle{icml2024}
\clearpage

% --------------- 初步的数据集中文草稿 ---------------
% 18847 个录音包含约 260K 个语音实例，平均语音长度为 500 个字符。
% 我们从以下几个纬度来描述数据集情况：
% - 投票信息：这里我们列举了 9 人狼人杀游戏中的角色每轮互投情况，如图 4 所示。我们可以看到，在人类玩家的数据分布中，76.0% 的预言家投票给狼人，15.7% 的狼人投票给队友，而其它好人角色大约有 50% 左右概率会投票给狼人。这和我们平时的游戏体验相同，预言家因为具有查验能力，能更确定地找出狼人，而狼人为了保护狼人阵营，会避开投票给狼队友。从狼人的投票偏好上来看，他们最喜欢投票给村民，概率大概是 28.6%，其次喜欢投票给预言家，概率大概是 28.1%，这和狼人喜欢诬陷视角有限的平民，以及希望尽快让预言家出局，避免预言家披露更多对自己不利信息的策略有关。
% - 胜率：如表 1 所示，在真实对局中，好人阵营和狼人阵营的胜率相近，分别为 49.31% 和 50.69%.
% - 发言时间：如图 5 所示，不同角色的发言持续时间略有差异，排名前三的角色分别是：预言家、村民、狼人，平均发言时长在 1min30s 左右。通常来说，预言家需要表明自己的查验信息、分析自己查验的心路历程，同时还要分析场上局势，赢得好人的信任，所以他需要有一定的说服力，发言时长最长。而狼人和村民需要尽可能多得说清自己视角下的角色预测，来证明自己的身份，所以也需要大量的发言。
% - 发言文本长度：如图 6 所示，在狼人杀中，不同角色的发言文本长度存在差异。我们发现预言家的发言长度最长，而狼人的发言长度最短，发言文本长度和图 5 中发言时间并不成正比例。狼人的发言时间较长，而发言文本较少，可能是由于狼人在欺骗时需要思考，体现了狼人发言难度。

% - 角色最终状态：如图 7 所示，它展示了人类对局中不同角色的最终状态，主要有存活、枪杀、毒杀、狼杀、票杀和自爆这几种类型。具体而言，女巫最容易被狼人夜晚杀害，比例在 55.3%，其次是预言家，比例为 32.5%；狼人自爆的概率为 17.3%，在夜晚死亡的场景下，狼人自刀概率最低，为 2.5%；在白天被票杀出局的角色分布中，概率最大的是狼人，最小的是女巫，说明对于白天的发言，狼人的欺骗难度最高，而女巫发言最容易，只需要说明客观信息就能赢得大家的信任。

% --------------- 初步的数据集中文草稿 ---------------

% \begin{enumerate}[label=(\arabic*)]
%     \item Vote information:
%     \item Win rate:
%     \item Speak Duration:
%     \item Speech Length:
%     \item Final State of the Roles:
% \end{enumerate}

\appendix

\onecolumn

\begin{figure*}[h]
\begin{center}
\includegraphics[width=1\textwidth]{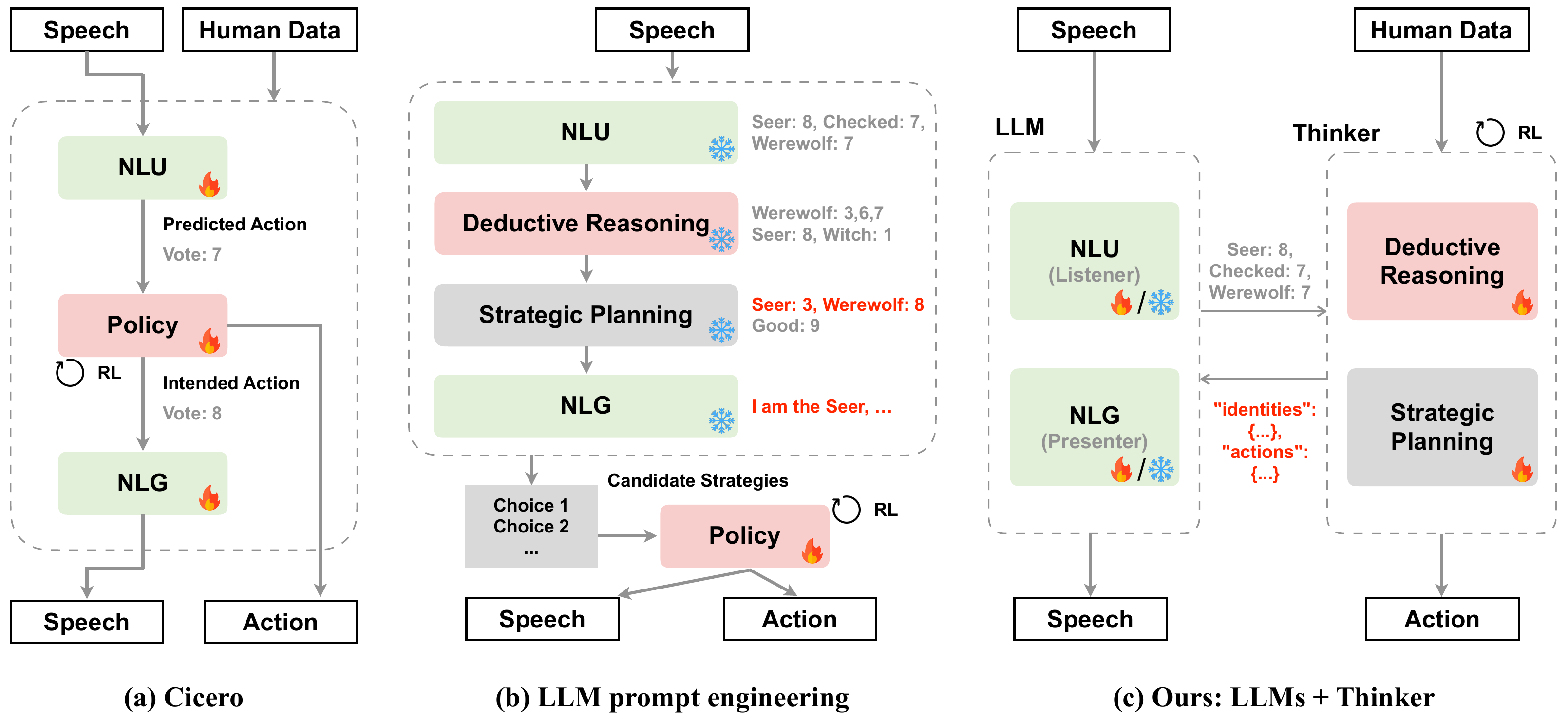}
\end{center}
\vspace{-10pt}
\caption{Comparing our framework with related approaches.} 
\label{fig:design_principal}
\end{figure*}

\section{Design Principal}
Figure~\ref{fig:design_principal} illustrate the comparison of Cicero approach~\cite{meta2022human}, LLM prompting-related approaches~\cite{xu2023language} and our proposed LLMs with \Thinker module.
We detail the evolving process of our framework as follows.

\subsection{Motivation}
In the game of werewolf, there is a significant gap between what a player says and what the player is actually thinking. 
Consider the scenario depicted in Figure~\ref{fig:thinker_induced_example}, where Player 3, a werewolf, publicly states: 

\emph{"I am the Seer, and I have checked Player 9, who is a good person. I suspect that Player 8 is a werewolf." }

While the surface meaning of this speech (\emph{System-1}) is straightforward, Player 3's internal thought (\emph{System-2}) process might be as follows:

\emph{"Players 6 and 7 are my fellow werewolves (as per the game rules, werewolves know each other's identities), and Player 8 claims to be the Seer and has accused Player 7, who is on my team. Therefore, Player 8 is likely the real Seer. By also pretending to be the Seer and verifying Player 9 as a villager, I can create a conflict with Player 8 in the eyes of the villagers."}

\subsection{LLM Prompting Methods}
We identified several shortcomings when examining the performance of LLM with typical prompt or mechanism engineering methods. 
The shortcomings are concluded into twofolds:

\textbf{Over-trust}: LLMs exhibited a tendency to over-trust other players' self-declared identities, particularly when players claimed to be Seer or Witch roles.
Furthermore, when the LLM assumed the role of a Werewolf itself, it was prone to inadvertently exposing its own identity, which is demonstrated in Section~\ref{sec_exp_nlg} and Table~\ref{tab_nlg}.

\textbf{Strategic Lackness}: LLMs showed a lack of familiarity with the common strategies employed in Werewolf games.
For instance, they failed to grasp tactics such as Werewolves pretending to be Seers to mislead other players, Werewolves accusing their teammates to gain the trust of the "Good" players, or Villagers pretending to be Seers to protect the real Seer from being killed.
These are conventional tactics used by experienced human players to navigate the complex social dynamics of Werewolf, which involve deception, trust, and betrayal.

To delve deep into the reasoning process of LLMs, we dissected the process from listening to speaking in the game into four stages, as shown in Figure~\ref{fig:design_principal} and investigate issues one by one:

\begin{enumerate}[label=(\arabic*)]
\setlength{\itemsep}{0pt}

\item \textbf{Natural language understanding}: It is assigned as
the Listener's goal in Figure~\ref{fig:werewolf_framework}, is to interpret speeches and extract their explicit meanings.
LLMs show proficiency in this area.

\item \textbf{Deductive reasoning}: LLMs underperform in role identification, often over-trust other players’ self-declared identities, as tested in Section~\ref{exp_nlu}.
Then the deductive reasoning is limited to information extraction. 

\item \textbf{Speech strategic planning}: LLMs struggle to outline a comprehensive speech plan, especially when assuming the role of a Werewolf.
They frequently risk exposing themselves or their allies (see Table~\ref{tab_nlg}), lacking an understanding of conventional Werewolf game speech strategies.

\item \textbf{Natural language generation}: Although LLMs are unfamiliar with conventional speech strategies, we find that they can generate sound and convincing speeches once prompted with basic instructions, e.g., "You should pretend to be the Seer, and accuse the Player 3 as a werewolf".

\end{enumerate}

\subsection{Transition to the \Thinker Module}
The primary reason for the above shortcomings is that LLMs are not trained on Werewolf-specific knowledge corpus and data.
Although it is possible to prompt LLMs with common game terminologies through in-context learning, strategic experiences are challenging to encapsulate in text prompts.
To address the deficiencies in deductive reasoning and speech strategic planning, we considered developing a trainable Thinker model to handle these aspects separately from the LLMs.
The Thinker module was optimized through imitation learning and reinforcement learning, using human game data as a foundation.
It was designed to complement the LLMs, which were responsible for more intuitive, System-1 reasoning tasks.

\subsection{Comparison with Cicero}

In brief, the differences between our approach and Cicero are as follows:

\textbf{Different roles for NLU and NLG}:
In Cicero's approach, both NLU and NLG involve a high-level logical reasoning process:
NLU directly outputs action predictions, which is actually a complex reasoning process that goes beyond natural language processing. 
Similarly, NLG takes intended actions as control signals, but it still requires a comprehensive consideration of the game state, historical speeches, and higher-level reasoning to generate reasonable dialogue/speech that matches the intended action.
In contrast, in our Werewolf game approach,
The Listener (NLU) is only responsible for extracting key information from speeches and does not infer the truthfulness of the speeches or the underlying intentions.
Similarly, NLG expands speech instructions, which are outlines of speeches, into full statements in context, requiring less  domain-specific reasoning.

\textbf{The connection between LLMs and policy}:
In Cicero's approach, the connection between LLMs and policy is made only through action prediction and intended action, which is non-language-based.
In the Werewolf game scenario, we found that using actions alone is not sufficient, as the Listener causes significant information loss. Due to the complexity of Werewolf speeches, intended actions also struggle to describe and control speech generation. This leads to a noticeable disadvantage for Cicero's approach in the ablation study presented in Table 4 and Table 5.
To address this, we propose a language-based feature and speech instruction that include complex verbal information, which can effectively summarize player speeches and control the speech generation process.

\textbf{Different training modes}:
Due to Cicero's method involving NLU and NLG in task-specific high-level reasoning processes, it is necessary to fine-tune both NLU and NLG.
In our approach, by defining explicit language-based connections and isolating domain-specific complex reasoning from LLMs with the Thinker, we can avoid the fine-tuning of NLU and NLG.

\section{Additional Results and Ablation Studies}
\label{sec_ablation}

\subsection{Legal Speak Generation} 
The ratio of legal speech generation from human evaluation is shown in Table~\ref{tab_nlg} with the criteria detailed in Appendix~\ref{eval_nlg}.
We can conclude that the speech instruction improve the legality of speeches for all the roles, especially when GPT4 playing the werewolf.

\begin{table*}[]
\centering
\caption{Legal speak generation ratio from human evaluation.}
\begin{tabular}{
p{2.0cm}
p{1.2cm}<{\centering}
p{1.2cm}<{\centering}
p{1.2cm}<{\centering}
p{1.2cm}<{\centering}
p{1.2cm}<{\centering}
p{1.2cm}<{\centering}
}
\toprule
Method & Total & Seer & Witch & Hunter & Villager & Werewolf \\
\midrule
GPT3.5-LtM & 68.0\%   & 90.6\% & 84.4\% & 81.8\% & 93.8\% & 24.8\% \\
GPT3.5-T & \textbf{75.4\%} & 96.9\% & 100\%  & 100\%  & 98.8\% & 28.6\% \\
\midrule
GPT4-LtM & 86.3\% & 100\%  & 90.6\% & 97\%   & 96.2\% & 66.7\% \\
GPT4-T & \textbf{98.7\%} & 100\%  & 100\%  & 100\%  & 100\%  & 96.2\% \\
\midrule
Finetune-T & 96.9\% & 90.6\% & 100\%  & 100\%  & 97.5\% & 96.2\% \\
\bottomrule
\end{tabular}
\label{tab_nlg}
\end{table*}

\subsection{Predicting Action as Language Feature}
\label{sec_cicero}
We study the approach used by Cicero~\cite{meta2022human}, utilizing the prediction of players' future actions as a feature representation of speeches and as a control variable for the speech generation.
Aside from example illustrated in Figure~\ref{fig:werewolf_framework}, we additionally conduct experiments by feeding the model with complete game states and historical speeches to predict players' future actions.
We fine-tune the ChatGLM-6B~\cite{du2021glm} model using data from the \fanlang dataset and then tested the action prediction accuracy on a set of 100 test games. 

The results are shown in Table~\ref{tab_cicero}.
Overall, the action prediction accuracies for three days do not exceed 40\%.
Notably, the Witch conventionally save the player killed by werewolves on the first day, resulting in a high accuracy. 
One point of particular interest is the accuracy of voting predictions, which consistently remained just over 40\% as the days progressed. 
In the game of Werewolf, the speaking order plays a crucial role; players who speak earlier often mention multiple potential voting targets.
By listening to subsequent speeches, players can make informed decisions or adjustments regarding their final vote. 
This aspect of the game dynamics makes the implementation of Cicero's method challenging in the context of Werewolf.

\begin{table*}[]
\centering
\renewcommand{\arraystretch}{1.1}
\caption{Accuracy of predicting future actions.}
\begin{tabular}{l|c|ccc|cc} 
\toprule
\multirow{2}{*}{Time} & \multirow{2}{*}{Total} & \multicolumn{3}{c|}{Night skills} &  \multicolumn{2}{c}{Day actions} \\
& & Werewolves & Witch & Seer &  Hunter & Vote \\
\midrule
Date1 & 
\makecell[c]{37.0\% \\ \small{[422/1142]}} & \makecell[c]{13.3\% \\ \small{[40/300]}} & \makecell[c]{97.0\% \\ \small{[97/100]}} & \makecell[c]{12.0\% \\ \small{[12/100]}} &\makecell[c]{ 0.0\% \\ \small{[0/4]}} & \makecell[c]{42.8\% \\ \small{[273/638]}} \\
\midrule
Date2 & \makecell[c]{30.3\% \\ \small{[268/884]}} & \makecell[c]{17.0\% \\ \small{[51/300]}} & \makecell[c]{20.6\% \\ \small{[20/97]}} & \makecell[c]{18.4\% \\ \small{[14/76]}} & \makecell[c]{10.0\% \\ \small{[1/10]}} & \makecell[c]{45.4\% \\ \small{[182/401]}} \\
\midrule
Date3+ & \makecell[c]{36.6\% \\ \small{[128/350]}} & \makecell[c]{34.4\% \\ \small{[67/195]}} & \makecell[c]{30.0\% \\ \small{[3/10]}} & \makecell[c]{22.7\% \\ \small{[5/22]}} & \makecell[c]{33.3\% \\ \small{[1/3]}} & \makecell[c]{43.3\% \\ \small{[52/120]}} \\
\bottomrule
\end{tabular}
\label{tab_cicero}
\end{table*}

\subsection{Comparison with Other Approaches}

In this section, we compare the performance of our proposed method, a Cicero-like baseline variant, and the approach described in ~\cite{xu2023language}.
To ensure a rigorous experimental comparison, we adapted the implementations of the comparative methods to account for differences in implementation details, thereby enhancing the persuasiveness of our results.
Below we outline the configurations for each method:

\textbf{Our Method}: We employ the \textbf{GPT4-T} setting, wherein the \Listener and \Presenter components utilize GPT4, and the \Thinker component is powered by the RL-optimized model.

\textbf{Variant of Cicero}: For this baseline, we reduce the language feature and speech instruction dimensions to a single dimension, representing the future action of a speaking player.
As experimental findings in Appendix~\ref{sec_cicero} indicated that fine-tuning ChatGLM-6B yielded low action prediction accuracy, we directly use GPT4 to generate language features and speech instructions in the \Listener and \Presenter. 
The \Thinker component employs an RL model for training, with its language feature and speech instruction also condensed to one dimension.
All other configurations are consistent with GPT4-T.

\textbf{Variant of~\cite{xu2023language}}: Diverging from the original implementation, we modify the approach to have GPT4 generate three speech instruction candidates instead of directly producing speak candidates.
The \Thinker then selects one speech instruction, which is subsequently used by the GPT4 \Presenter to generate speech.
Due to the discrepancy between LLM inference and \Thinker RL sampling speeds, the \Thinker is restricted to using offline RL. 
For offline RL data construction, we extracted \num{1000} game sessions from the \fanlang dataset.
For each instance of speaking, we allow GPT generate five speech instruction candidates.
During offline RL training, we randomly selected two of the five GPT-generated candidates and combined them with the human speech instruction to form three speech instruction candidates, yielding \num{10} possibilities for data augmentation.
The \Thinker makes its selection, with its inputs including the game state, language features as in GPT4-T, and the three speech instruction candidates. 
The actual selection for BC is the human speech instruction.

To summarize, the primary distinction between GPT4-T and the Cicero variant lies in the modification of the dimensions and meanings for language feature and speech instruction. 
And the \Thinker in the variant of~\cite{xu2023language} no longer generates speech instructions; instead, it directly selects from generated candidates.
The evaluation results are shown in Table~\ref{tab_ablation}.
Our GPT4-T method surpasses the variant of~\cite{xu2023language} in performance, and significantly outperforms the Cicero variant, highlighting the advantages of external \Thinker module in terms of reasoning and strategic communication within the Werewolf game.

\begin{table}[h]
\centering
\caption{Win rate comparison of our method with other approaches.}
\begin{tabular}{
p{6cm}
p{1.5cm}<{\centering}
p{1.5cm}<{\centering}
p{1.5cm}<{\centering}
% p{2.1cm}<{\centering}
}
\toprule
Method & Total & Goods & Werewolves \\
\midrule
Variant of Cicero~\cite{meta2022human} & 34.4\% & 28.5\% & 47.9\% \\
Variant of~\cite{xu2023language}  & 47.8\% & 37.4\% & 67.7\% \\
Ours (GPT4-T)     & \textbf{53.5}\% & \textbf{41.6}\% & \textbf{75.2}\% \\

\bottomrule
\end{tabular}
\label{tab_ablation}
\end{table}

\subsection{Training Curve}
\begin{figure}[h]
\begin{center}
\includegraphics[width=1.0\linewidth]{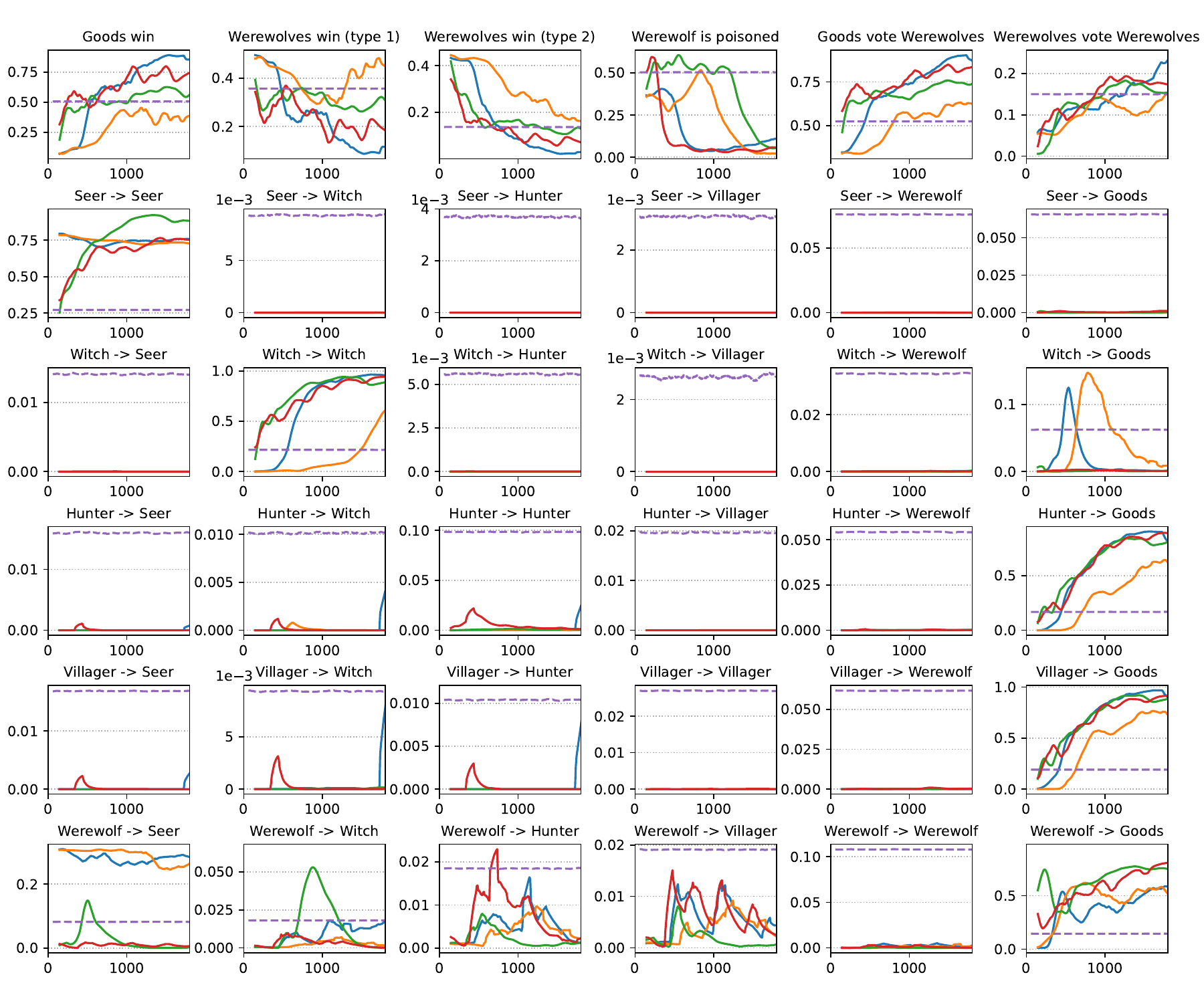}
\end{center}
\vspace{-0.5cm}
\caption{
Detailed training curves for different agents during RL training.
The x-axis represents the training steps (k), and the y-axis represents the probability.
The horizontal line in each subplot corresponds to the probability observed in human data.
"Werewolf -> Seer" represents that a Werewolf claims that he is the Seer in the speech.
}
\label{fig:population}
\end{figure}

The population-based RL training of different agents is illustrated in Figure~\ref{fig:population}.

\vspace{-5pt}

\section{Game Rules}
\label{sec:gamerule}
We follow the 9-player standard mode Werewolf game rules on the Fanlang platform.
The rules are outlined as follows.

\subsection{Objectives}
The game is divided into two factions: the "Good" faction, which includes Villagers and special roles, and the "Werewolf" faction.
Additionally, there is a Moderator who is responsible for managing the game and ensuring the rules are followed.
The goal for the "Good" faction is to identify and execute all Werewolves, while the goal for Werewolves is to kill or exile all Villagers or all special roles.
The game ends when any of the following conditions are met:
\begin{itemize}[leftmargin=*]
\setlength{\itemsep}{0pt}
    \item All Villagers are out of the game (Werewolves win)
    \item All special roles are out of the game (Werewolves win)
    \item All Werewolves are out of the game ("Good" faction win)
\end{itemize}

\subsection{Roles}
The game comprises 3 Villagers, 3 Werewolves, and 3 special roles (Seer, Witch, and Hunter).
The identities of the players are hidden from each other, even after being eliminated from the game.

\textbf{Werewolves}: Werewolves are aware of each other's identities.
At night, they decide to kill a living player, which can include themselves. 
The majority of the Werewolves' choice will be the final kill target. 
If there is a tie, a random player in the tie is killed. 
Werewolves can commit suicide during the speech sessions, which will reveal their identity, and the game immediately proceeds to the night phase, skipping the remaining daytime processes such as speeches and voting.

\textbf{Villagers}: Villagers have no special abilities. 
They must determine other players' identities based on their speeches and vote to exile potential Werewolves.

\textbf{Seer}: The Seer can verify a player's faction each night (a Werewolf or the "Good"), but cannot know their specific role. 
The Seer cannot verify himself or any player who has already been verified.

\textbf{Witch}: The Witch possesses an antidote and a poison. 
The antidote can save a player killed by Werewolves at night, and the poison can kill a player.
The Witch cannot use both potions in the same night and can only save herself on the first night.

\textbf{Hunter}: When the Hunter is killed by Werewolves at night or voted out during the day, he can shoot a player. 
However, the Hunter cannot use his ability when poisoned by the Witch.

\subsection{Game Task Flow}

The game proceeds in a night-day cycle until the victory conditions are met.

\textbf{The night tasks flow}:
\begin{enumerate}[label=(\arabic*)]
\setlength{\itemsep}{0pt}
    \item Werewolves decide to kill a player.
    In our simulation of the game environment, we have simplified the discussion into a three-round voting process.
    During voting, werewolf players can see their teammates' previous votes.

    \item The Witch uses her ability.
    \item The Seer uses his ability.
\end{enumerate}
\textbf{The daytime tasks flow}:
\begin{enumerate}[label=(\arabic*)]
\setlength{\itemsep}{0pt}

    \item The Moderator announces the deaths from last night but does not reveal the causes of death.
    \item Deceased players give their last words (only for the first day).
    \item If deceased players have additional abilities, they may choose to use them.
    
    \item First round of speeches. 
    The speech sequence is determined by the following rules:
    (a) if no player died last night, randomly select an initial speaker and randomly decide a clockwise or counterclockwise speaking order.
    (b) randomly select a deceased player and start the speaking order clockwise or counterclockwise from him.
    Players cannot interrupt others during their speeches.

    \item First round of voting. 
    Each player votes for a single player to exile from the game.
    Other players' voting choices remain hidden until the voting session ends.
 
    \item Second round of speeches.
    If there is a tie in the first round of voting, the tied players give their second speeches; otherwise, the process moves on to task (8)   
    The first speaker, selected randomly from the tied players, initiates the sequence, which could proceed either clockwise or counterclockwise.
    \item Second round of voting. If there is still a tie after the second vote, the game moves on to the next night, and no player is exiled.
    \item The exiled player gives his last words.
    \item If exiled players have additional abilities, they may choose to use them.
\end{enumerate}

% \textbf{Additional Rules:}
% To provide a good game experience for all players, there are some conventional rules:
% \begin{enumerate}[label=(\arabic*)]
%     \item The Witch must save the player killed by Werewolves on the first night.
%     \item The Witch cannot use the poison on the first night.
% \end{enumerate}

\begin{figure*}[t]
\begin{center}
\includegraphics[width=0.8\textwidth]{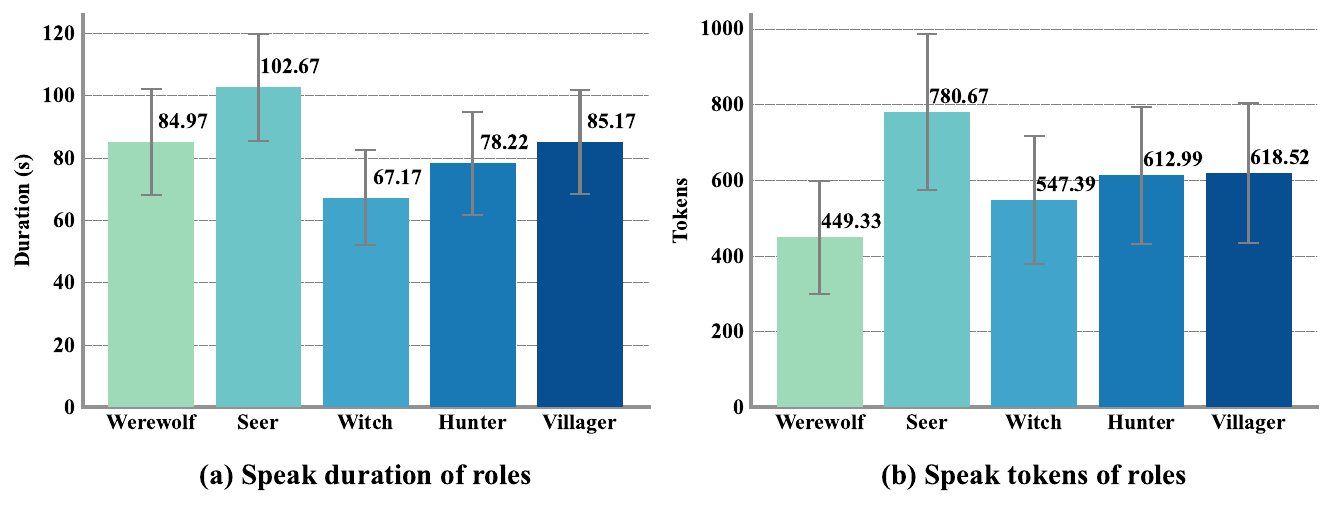}
\end{center}
\vspace{-10pt}
\caption{Speech duration and token length categorized by roles in \fanlang dataset.} 
\label{fig:combined_duration_and_tokens}
\end{figure*}

\begin{figure}[h]
\begin{center}
\includegraphics[width=.6\linewidth]{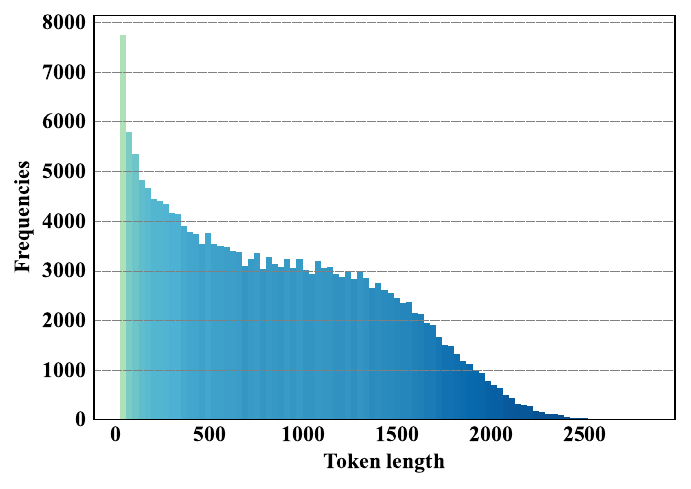}
\end{center}
\vspace{-0.5cm}
\caption{Distribution of speech token length.} 
\label{fig:speech_tokens_distribution}
\end{figure}
\vspace{-5pt}

\begin{figure*}[t]
\begin{center}
\includegraphics[width=1.0\textwidth]{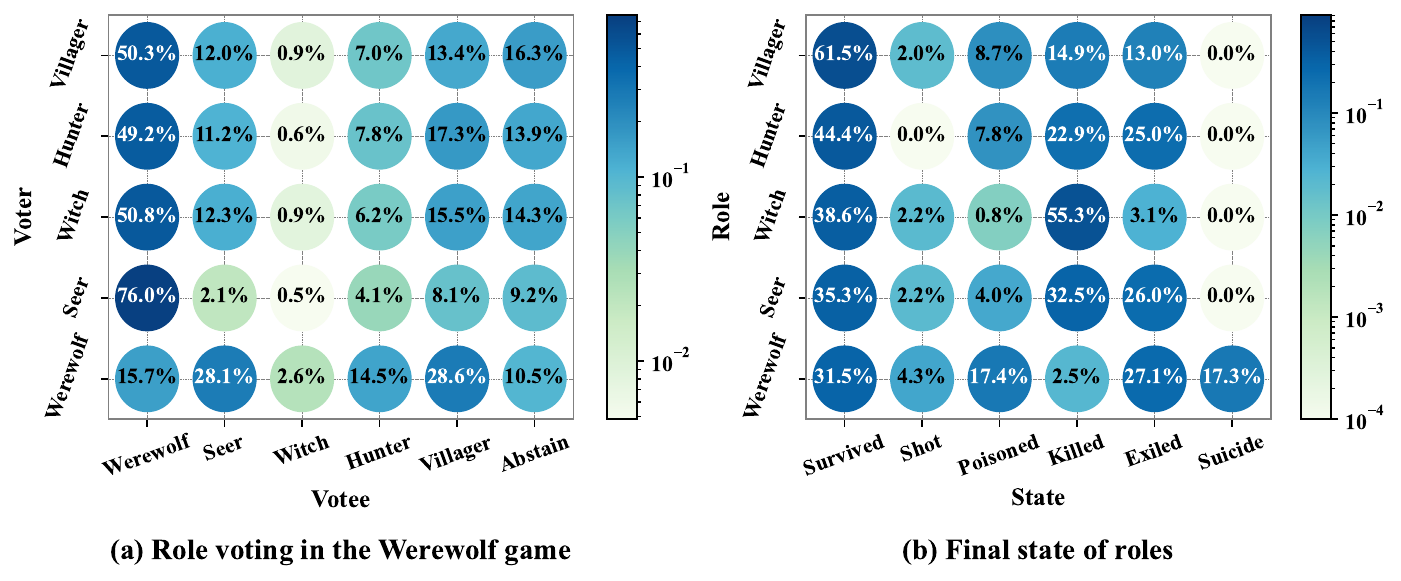}
\end{center}
\vspace{-10pt}
\caption{(a) The voting probability distributions for players with different identities in all voting sessions; (b) the final survival status and causes of death probabilities for players at the game's end.}
\label{fig:combined_vote_and_state}
\end{figure*}

\section{Analysis of the \fanlang Dataset}
\label{sec:fanlang9}
The \fanlang dataset consists of \num{18800} recordings, \num{260}K speech instances, with an average speech length of \num{500} characters.
Specifically, the following characteristics underscore the unique nature of the dataset:

\subsection{Speech Duration and Length}
Figure~\ref{fig:combined_duration_and_tokens} (a) demonstrates significant variations in speech duration among different roles, with an average of approximately \num{90} seconds each. 
The Seer's inspection information at night forms the core and fundamental logical basis of the game.
Therefore, it is the Seer's duty to share inspection information, provide persuasive speeches, and lead discussions during the speech phase, resulting in the longest duration among all roles.
Besides, Werewolves and Villagers need to convincingly identify themselves and predict the roles of other players, necessitating detailed and logical analysis. 
In Figure~\ref{fig:combined_duration_and_tokens} (b), the dataset shows the shortest token length among Werewolves, which is not correlated with their speaking time. 
This suggests that Werewolves' speeches are relatively concise, which may stem from the complexity of deception that requires more time to strategize.
We further illustrate the distribution of token length in a single speech in Figure~\ref{fig:speech_tokens_distribution}.

\subsection{Tokenization and Categorization of Speeches}
The reasoning result of a speech produced by the \Listener is formatted in JSON style, containing pairs of player ids with their attributes.
The result typically includes phrases and word groups containing multiple attributes, probabilities, and irrelevant information, e.g., "seems to be a werewolf: [3, 6]", "cannot hear clearly: [8]".
We then tokenize and categorize the result into related identities and actions, along with their probabilities, as shown in Table~\ref{table:word_frequency}.
The final language features account for 96.09\% of the \fanlang dataset, capturing the majority of the information expressed by speakers in the Werewolf game.

\subsection{Voting Preference}
We analyze how human players tend to vote in the perspective of different roles in Figure~\ref{fig:combined_vote_and_state} (a).
As for voting werewolves, the Seer has the highest accuracy of voting Werewolves due to his inspection ability, while Werewolves vote for their teammates with a probability of 15.7\%, aiming to disguise themselves as the "Good" faction.
The other roles have a 50\% chance of voting for Werewolves, since they lack additional information beyond the game state and historical speeches.
As for voting from Werewolves, the most prioritized target are the Villagers (28.6\%), since they have the least amount of information and are easier to be incriminated as Werewolves.
The second prioritized target is the Seer (28.1\%), since the Seer can inspect players' identities, it is crucial to remove him out of the game as soon as possible.

\subsection{Final State of the Roles}
In Figure~\ref{fig:combined_vote_and_state} (b), we present the final states of roles in the end of the game, categorized as \emph{Survived}, \emph{Shot} by the Hunter, \emph{Poisoned} by the Witch, \emph{Killed} by Werewolves, \emph{Exiled} after the Voting stage, and Werewolves committed \emph{Suicide}.
Notably, the Witch has the highest likelihood of being killed by the werewolf at night (55.3\%), with the seer following at 32.5\%.
Werewolves commit suicide with a probability of 17.3\%, and are killed by their teammates at night with a probability of 2.5\%.
During the daytime voting, Werewolves are the most frequently exiled role, indicating their challenges in providing deceptive statements, while the Witch has the lowest probability, reflecting their effectiveness in gaining trust through speeches.

\subsection{Win Rate}
Table~\ref{table:win_rate} illustrates that in human gameplay, the win rates for the Good and Werewolf factions are closely matched.

\begin{table}[h]
\begin{center}
\renewcommand\arraystretch{1.1}
\setlength{\tabcolsep}{4mm}{}
\caption{Win rate in the \fanlang dataset.}
\label{table:win_rate}
\begin{tabular}{lcc}
\toprule
Camp & Win number & Win rate \\
\midrule
Goods & 9293 & 49.31\% \\
Werewelf & 9554 & 50.69\% \\
\bottomrule
\end{tabular}
\end{center}
\end{table}

\begin{table*}[h]  
\centering  
\caption{Tokenization and categorization of speeches on the \fanlang dataset.}
\begin{tabular}{l|ccccc|cc}  
\toprule  
Tokenized attributes & Is & Might be & Is not & Might not be & Is not sure & Ratio & Accumulation\\   
\midrule  
Werewolf & \num{178423} & \num{27297} & \num{516} & \num{313} & \num{15} & \num{26.55}\% & \num{26.55}\% \\  
Good (the good faction) & \num{83071} & \num{622} & \num{85} & \num{73} & \num{10} & \num{10.77}\% & \num{37.32}\%  \\  
Vote & \num{68853} & \num{87} & \num{81} & \num{1} & \num{3}  & \num{8.87}\% & \num{46.19}\%  \\  
Seer & \num{60339} & \num{114} & \num{111} & \num{321} & \num{8}  & \num{7.82}\% & \num{54.01}\%  \\   
Witch & \num{35408} & \num{42} & \num{29} & \num{8} & \num{3}  & \num{4.56}\% & \num{58.57}\%  \\   
Gold Water (checked Good) & \num{34727} & \num{8} & \num{8} & \num{1} & /  & \num{4.46}\% & \num{63.03}\%  \\   
Check (Seer's inspection) & \num{26027} & \num{17} & \num{17} & / & /  & \num{3.35}\% & \num{66.38}\%  \\  
Poison & \num{21897} & \num{82} & \num{9} & \num{1} & /  & \num{2.83}\% & \num{69.21}\%  \\  
Villager & \num{21611} & \num{28} & \num{19} & \num{10} & \num{1}  & \num{2.78}\% & \num{71.99}\%  \\  
Werewolves' target & \num{19481} & \num{17} & \num{12} & / & \num{1}  & \num{2.51}\% & \num{74.50}\%  \\  
Hunter & \num{17603} & \num{26} & \num{70} & \num{5} & \num{2}  & \num{2.28}\% & \num{76.78}\%  \\    
Silver Water (saved) & \num{14016} & \num{3} & \num{5} & \num{1} & \num{2}  & \num{1.80}\% & \num{78.58}\%  \\    
Suicide & \num{3826} & \num{4} & \num{1} & / & \num{1} & \num{0.49}\% & \num{79.07}\%  \\    
Uncertain Identity & / & / & / & / & \num{2937} & \num{0.38}\% & \num{79.45}\%  \\    
Shoot & \num{1100} & \num{2} & \num{2} & / & /  & \num{0.14}\% & \num{79.59}\%  \\    
Save (by the Witch) & \num{1065} & / & / & / & / & \num{0.14}\% & \num{79.73}\%  \\    
Abstain voting & \num{683} & \num{3} & \num{1} & / & /  & \num{0.09}\% & \num{79.82}\%  \\     
Special Role & \num{273} & \num{4} & / & / & /  & \num{0.04}\% & \num{79.86}\%  \\    
\midrule    
Irrelevant Information & \num{126279} & / & / & / & /  & \num{16.23}\% & \num{96.09}\%  \\  
Unprocessed & \num{30476} & / & / & / & / & \num{3.91}\%  & \num{100.00}\% \\
\bottomrule
\end{tabular}
\label{table:word_frequency}
\end{table*}

\section{Ethical Considerations}
With the integration of LLMs into complex reasoning tasks, such as those demonstrated in social deduction games like Werewolf, we are witnessing the emergence of AI agents that not only mimic human-like reasoning but also engage in communications that could be considered deceptive by nature. 
While these developments showcase the potential of AI to understand and navigate intricate human interactions, they also raise important ethical and societal considerations that must be addressed.
To address these ethical and societal challenges, we propose several mitigation strategies:

\textbf{Transparent communication and monitoring}: 
Our framework ensures transparency through explicit structured information at every stage of the AI's decision-making process, from listening and reasoning to speech generation. To enhance this transparency, we propose implementing real-time transparency logs that capture and display the reasoning paths, identity predictions, and speech instructions generated by the AI. 
By having a complete audit trail, we can monitor the AI's decision processes, ensure adherence to ethical guidelines, and trace any unintended actions back to their source.

\textbf{Control and filtering mechanisms}: 
Our speech instructions are enriched with contextual information specific to the Werewolf game, allowing for robust control over the fine-tuned LLM. 
To further mitigate potential negative impacts, we propose implementing dynamic contextual guardrails.
These guardrails will utilize our existing filtering mechanism (as outlined in Section~\ref{sec:presenter}) to not only match generated speech with instructions but also to check against a set of ethical and societal norms. If the AI's output is flagged as potentially harmful or deceptive beyond the game's scope, it will be withheld and replaced with a template response. This additional layer of control will act as a safeguard against the misuse of AI in generating deceptive or manipulative content outside the intended gaming environment.

\section{Implementation Detail}
\label{sec_implementation}

\subsection{Evaluation Criteria for the Speech Generation}
\label{eval_nlg}
The human evaluation requirements for speech generation are as follows. 

\textbf{Legality:} Absence of obvious logical errors and illegal statements that are conflicted with game rules, such as:
\begin{itemize}[leftmargin=*]
\setlength{\itemsep}{0pt}
\item "I am a Werewolf." 
\item "I am the Seer, and I poisoned Player 5 last night."
\item "Player 3 is a good person; I suggest voting for him."
\item "I suggest voting for myself."
\item "Player 8 is a Werewolf, he was voted out and took Player 6." (Player 8 is the hunter and publicly shot Player 6).
\item "I suggest voting for Player 8." (Player 8 has already been voted out).
\end{itemize}

\textbf{Reasonableness:} of the speeches, such as
\begin{itemize}[leftmargin=*]
\setlength{\itemsep}{0pt}
\item The Seer correctly reports his inspection last night.
\item Werewolves reasonably disguise their identity, employing various strategies such as pretending to be the Seer, aggressive claims, and betraying their teammates.
\item Villagers make reasonable guesses about the Good faction and Werewolves. 
\item Note: the correctness of guessing other players' identities is not part of the evaluation criteria.

\end{itemize}

\textbf{Other:} factors unrelated to key information:
\begin{itemize}[leftmargin=*]
\setlength{\itemsep}{0pt}
\item Language style, colloquial expression, game jargon. 
\item Presence of verbose or redundant statements, such as greetings or defending the village community.
\end{itemize}

The evaluation criteria are in descending order of priority.
For example, if model A has no obvious logical errors but its speech is not very reasonable, and model B has obvious logical errors, then A is better than B.
For the ranking of the five samples, if there are obvious logical errors, mark them as $-1$ and no need to rank them. For example, if models A and B have obvious errors, the annotation result could be: 
${\rm{A}: -1, \rm{B}: -1, \rm{C}: 1, \rm{D}: 2, \rm{E}: 3}$,
where $1$ represents the best and $5$ represents the worst. Apart from marking illegal statements as $-1$, tied rankings are not allowed.

\begin{table*}[htbp]  
\centering  
\caption{Behavior scores applied in the 9-player werewolf game.}  
\begin{tabular}{llc}  
\toprule  
Role & Description & Score \\
\midrule
\multirow{2}{*}{Seer} & If a werewolf is exiled in the first day & $+0.5$  \\
                      & For giving up the inspection at night & $-0.5$ \\
\midrule

\multirow{2}{*}{Witch} & For poisoning a werewolf & $+1.0$  \\
                       & For poisoning a good player & $-1.0$ \\
\midrule
\multirow{2}{*}{Hunter} & For shooting a werewolf & $+1.0$ \\ 
                        & For shooting a good player  & $-1.0$ \\
\midrule
\multirow{2}{*}{Good roles except the Seer} & For voting for a werewolf & $+0.5$ \\
                                       & For voting for a good player & $-0.5$ \\
\bottomrule 
\end{tabular}
\label{tab:behavior_score}  
\end{table*}

\subsection{\Thinker Model Structure}
The architecture of the \Thinker network is designed to capture the intricacies of gameplay from the perspective of the current player, which encompasses speeches, actions, and game status information of all nine players involved, including themselves.
We employ a shared-parameter feature encoding network that processes the data for each of the nine players individually.

For the $i$-th player, up to \num{10} language features $\mathbf{F}$ are stored.
These language features are enriched with headers indicating the time-tag, type, and order of the speeches. Subsequently, these annotated language features are processed through another shared-parameter speech feature encoding network, which consists of a three-layer (181-256-256) multilayer perceptron network (MLP). 
After processing the ten pieces of features, a \emph{reduce\_mean} operation is applied to the outputs to synthesize the overall speech embedding for the player $e_{i}^{\rm{speech}}$.
This synthesized speech embedding is then combined with additional game state information such as the player's actions, status, and other relevant data.
The aggregated data is fed through a feature encoding network (again, a three-layer MLP of 1019-512-512) to generate the feature embedding for the $i$-th player $e_{i}$.

In the final step, the feature embeddings of all nine players $e_{1},e_{2},...,e_{9}$ are subjected to a \emph{reduce\_mean} operation to create a collective feature encoding.
This comprehensive encoding is then passed through an all-players feature encoding network (a three-layer MLP of 523-512-512) to construct the corresponding action decision, identity prediction headers, as well as speech instructions.

\subsection{Reward Shaping}
Drawing inspiration from the concept of the Behavior Score, we have devised the reward shaping for \Thinker in the  reinforcement learning to circumvent illegal actions and speech that may arise during unfettered exploration within the AI Werewolf game.
The specifics of this mechanism are outlined in Table~\ref{tab_reward}. 
It encompasses several key areas:

\begin{itemize}[leftmargin=*]
\setlength{\itemsep}{0pt}
\item Game result reward: The AI receives a reward based on the win or loss, survival duration at the end of a game.

\item Action reward: for taking actions that are deemed appropriate and effective within the context of the game.

\item Speech reward: to incentive the AI to engage in communication that is beneficial to its goals, such as persuading other players or disseminating useful information.
\item Action-Speech consistency reward: to stimulate coherence between what the AI says and does, a reward is given for alignment between the AI's declared intentions in speech and its subsequent actions.
\item Cognitive reward for Werewolves: Central to the training of a Werewolf AI is the ability to masquerade as a member of the "Good" faction.
To enhance this capability, we provide a reward based on the change in identity prediction from the perspective of the "Good" players. 
The better a Werewolf AI can deceive the "Good" faction about its true identity, the larger the reward it receives.
\end{itemize}

\begin{table*}[h]
\begin{center}
\renewcommand\arraystretch{1.1}
\caption{
Reward shaping in the RL training of the \Thinker.
}
\label{tab_reward}
\begin{tabular}{lc}
\toprule
Description & Reward \\
\midrule
\multicolumn{2}{l}{\# Game reward}            \\
the Good faction win, Werewolves get & $-4$ \\
the Good faction win, Villagers and special roles get & $+2$ \\
Werewolves win, Werewolves get & $+4$ \\
Werewolves win, Villagers and special roles get & $-2$ \\
Any player survives for a new day & $+1$ \\
\midrule
\multicolumn{2}{l}{\# Action reward}            \\
the Goods vote for a Werewolf & $+2$ \\
the Goods vote for a Good role & $-2$ \\
the Witch poisons a Werewolf & $+2$ \\
the Witch poisons a Good role & $-4$ \\
the Hunter shoots a Werewolf & $+2$ \\
the Hunter shoots a Good role & $-4$ \\
\midrule
\multicolumn{2}{l}{\# Speak reward}            \\
the Seer claims his identity & $+2$ \\
the Witch claims his identity & $+1$ \\
the Goods correctly identify a Werewolf in the speech & $+2$ \\
the Goods wrongly identify a Werewolf in the speech & $-2$ \\
the Goods correctly identify a Good role in the speech & $+1$ \\
the Goods wrongly identify a Good role in the speech & $-1$ \\
Any player who claims that he is a Good role & $+0.5$ \\

\midrule
\multicolumn{2}{l}{\# Action-Speech correlated reward}            \\
the Seer correctly share his inspection last night & $+2$ \\
the Witch correctly share the usage of antidote or poison & $+1$ \\
any player who claims the voting intention and then vote the same player &  $+1$ \\
\midrule
\multicolumn{2}{l}{\# Cognition reward}            \\
\multicolumn{2}{l}{the change $\delta$ of summation of a Werewolf's identity probabilities in the Goods' perspective:}  \\

as the Seer & $4\delta$ \\
as the Witch & $2\delta$ \\
as the Hunter or Villagers & $1\delta$ \\

\bottomrule
\end{tabular}
\end{center}
\end{table*}

\subsection{Details of Overall Training Process}
\label{appendix-Pseudocode}
We provide Pseudo-code in Algorithm~\ref{alg_ppo}, the \Thinker and LLMs are trained separately in our framework.
This design choice was intentional and serves as one of the strengths of our framework.
The separation facilitates training efficiency since LLMs, which we employ as both \Listener and \Presenter, are inherently slower in sample generation compared to the \Thinker module. 
Therefore, to optimize our training process, we either employ offline RL or decouple the training of the \Thinker and LLMs.
The inference workflow is as follows:
\Listener (LLM) -> language feature $\mathbf{F}$ -> \Thinker (RL) -> speech instruction $\mathbf{I}$ -> \Presenter (LLM)

During the training of the \Thinker, the generated speech instructions $\mathbf{I}$ are treated as the new input language features $\mathbf{F}$ for the subsequent steps, allowing for a seamless integration of the RL training into the overall process.
Our hybrid training framework incorporates both BC and PPO. During training, each game session has a certain probability of being a BC or RL game. 
In a BC session, actions $a$ and speaking instructions $\mathbf{I}$ are taken directly from human replay, bypassing the \Thinker inference. Conversely, in an RL session, the \Thinker actively generates actions and speaking instructions.
Samples from the game session are tagged as either BC or RL. For the Learner, BC samples utilize the BC loss mentioned in Equation~\ref{eqn_bc}, while RL samples employ the PPO loss Equation~\ref{eqn_ppo}.

\begin{algorithm}[!htbp]
\caption{Pseudo-code for the overall training process.}

\textbf{Require:}
\begin{itemize}
\setlength{\itemsep}{0pt}
\setlength{\parsep}{0pt}
\setlength{\parskip}{0pt}
    \item Data pairs 1: for finetuning of the \Listener
    
    \textbf{Input}: [game state $s$, historical speeches $\mathcal{H}$, current player's speech $\mathbf{S}$]
    
    \textbf{Output}: [language feature $\mathbf{F}$]
    
    \item Data pairs 2: for finetuning of the \Presenter

    \textbf{Input}: [game state $s$, historical speeches $\mathcal{H}$, speech instruction $\mathbf{I}$]
    
    \textbf{Output}: [current player's speech $\mathbf{S}$]
    
    \item Data pairs 3: for behavioral cloning of the \Thinker
    
    \textbf{Input}: [game state $s$, historical collection of all language features $\mathcal{F}$]

    \textbf{Output}: [action $a$], or [speech instruction $\mathbf{I}$], decided by the current task type.
\end{itemize}

\hrulefill

\textbf{\Listener and \Presenter}: 

\If{use APIs}{
    \Listener: Use API for generating language features $\mathbf{F}$.
    
    \Presenter: Use API for generating speeches $\mathbf{S}$.  
}  
\Else{
    \Listener: Finetune model with Data pairs 1 and hyperparameters in Table~\ref{tab:llm_hyperparameters_listener}.
    
    \Presenter: Finetune model with Data pairs 2 and hyperparameters in Table~\ref{tab:llm_hyperparameters_listener}.
} 

\hrulefill

\textbf{\Thinker}:

Initialize network parameters for a population of $P$ agents:
$\{\theta_{1}, \theta_{2}, ..., \theta_{P}\}$.

Start multiple actors and learners in parallel. 

\textbf{Actors}:
\While {true}{
    Fetch the latest model from the learners.
    Add the latest checkpoint into a checkpoint list.
    
    Sample $N-1$ checkpoints from the list and the latest checkpoint.

    Decide the game episode is BC or RL, run an $N$-player game episode.

    \If{game episode is BC}{
        Get behavioral cloning training samples from Data pairs 3.
    }  
    \Else{
        Generate RL training samples.
    }  
    
    Accumulate samples in the form $x = (s, \mathcal{F}, a, \mathbf{I}, r, \rm{is\_BC})$
    and send them to the replay buffer.
}

\textbf{Learners}: 

\For{\(t \in {1,2,3,...}\)}{
    
    \For{\(p \in {1,2,...,P}\)}{
        Fetch a batch of samples for agent $p$ from the replay buffer.
    
        Calculate value loss and policy loss according to PPO algorithm in Equation~\ref{eqn_ppo}.

        Calculate behavioral cloning loss according to Equation~\ref{eqn_bc}.

        Calculate loss for auxiliary tasks.
       
        Update parameters $\theta_{p}$ using gradients on loss in Equation~\ref{eqn_all}.
    }
}

\label{alg_ppo}
\end{algorithm}

\subsection{Training Hyper-parameters}
\label{sec:training_hyper_parameters}
The training hyper-parameters for the \Thinker are provided in Table~\ref{tab_hyper_thinker}.

\textbf{Regarding the hyperparameters in Equation~\ref{eqn_all}}:
The Behavioral Cloning coefficient $\alpha$ determines the extent to which the RL policy refers to human strategies versus greedily selecting the RL strategy. 
We observed that when $\alpha$ decays to $0$, werewolves completely abandon the strategy of claiming to be the Seer, because the difficulty for werewolves to pretend to be the Seer is high, and it is relatively challenging for RL to optimize. 
A more favorable choice is to masquerade as a villager.
Therefore, we still maintain a small $\alpha = 0.01$ during the later stages of training.
As for the auxiliary task coefficient $\beta$, we tested values in $\{1.0, 0.1, 0.01\}$, and found that they had minimal impact on RL, as it is an auxiliary learning task.

The fine-tuning hyper-parameters for the \Listener and \Presenter are provided in Table~\ref{tab:llm_hyperparameters_listener}.

\begin{table*}[h]
\begin{center}
\renewcommand\arraystretch{1.1}
\caption{
Hyperparameters for the \Thinker training.
}
\label{tab_hyper_thinker}
\begin{tabular}{ll}
\toprule
Hyperparameters & Value \\
\midrule
Population size & 4 \\
Number of actors & 700 (CPUs) \\
Number of learners & 8 (GPUs) \\
Replay buffer size & 100k \\
Mini-batch size & 2048 \\
Optimizer & Adam      \\
Learning rate  & 2e-4   \\
Discount factor ($\gamma$) & 1.0 \\
GAE parameter ($\lambda$) & 0.9 \\
PPO clipping ratio & 0.2 \\
Value function coefficient $c_{1}$ & 0.5 \\
Entropy coefficient $c_{2}$ & 0.05 \\

Behavioral Cloning coefficient $\alpha$ & $0.1 \to 0.01$ \\
Auxiliary task coefficient $\beta$ & 0.1 \\
\bottomrule
\end{tabular}
\end{center}
\end{table*}

\begin{table*}[h]
\centering
\renewcommand\arraystretch{1.1}
\caption{Hyperparameters for fine-tuning the \Listener and \Presenter.}
\label{tab:llm_hyperparameters_listener}
\begin{tabular}{p{0.3\textwidth}p{0.1\textwidth}p{0.1\textwidth}}
\toprule
Parameter             & \Listener &   \Presenter  \\
\midrule
\multicolumn{3}{l}{\# Basic Training Parameters}            \\
Learning rate                   & 1e-4       &   1e-4    \\
Sequence length                 & 4096       &   8192    \\
Optimizer                       & {AdamW}    &   AdamW      \\
Adam beta1                      & 0.9        &    0.9   \\
Adam beta2                      & 0.999      &    0.999  \\
Adam epsilon                    & 1e-8       &    1e-8   \\
Train batch size                & 32         &    8    \\
Train epochs                    & 3          &    3   \\
Max steps                       & 5000       &    10000  \\
Warmup steps                    & 500        &    1000   \\
Max grad norm                   & 1.0        &    1.0   \\
\midrule
\multicolumn{3}{l}{\# Model Configuration}                 \\
Hidden size                     & \multicolumn{2}{c}{4096}              \\
KV channels                     & \multicolumn{2}{c}{128}               \\
Num layers                      & \multicolumn{2}{c}{28}                \\
Num attention heads             & \multicolumn{2}{c}{32}                \\
Layer norm epsilon              & \multicolumn{2}{c}{1e-5}             \\
Torch dtype                     & \multicolumn{2}{c}{{float16}}           \\
\midrule
\multicolumn{3}{l}{\# Distributed Training Settings}        \\
TP size                         &\multicolumn{2}{c}{ 2 }                \\
PP size                         & \multicolumn{2}{c}{1}                 \\
\midrule
\multicolumn{3}{l}{\# Attention Mechanism Configuration}    \\
Multi query attention           & \multicolumn{2}{c}{{True}}            \\
Multi query group num           & \multicolumn{2}{c}{2 }                \\
\bottomrule
\end{tabular}
\end{table*}

\subsection{LLM Prompting for \Listener and \Presenter}
\label{sec:llm_prompting_for_nlu}
The information extraction prompt for the \Listener module contains the following parts: 
\begin{itemize}[leftmargin=*]
\setlength{\itemsep}{0pt}
\item Description of the background of the Werewolf game, as shown in Table~\ref{box:werewolf_game_background_prompt}, which provides the game configuration, game rules, terminology, and descriptions of roles' identities and skills.

\item  Task requirements, as shown in Table~\ref{box:dialogue_understanding_requirements_prompt}.
The prompt describes the structured information in JSON format that we expect LLMs to produce, and we describe the appropriate values for each position of the structured command and limit the output within a reasonable range.

\item Few-Shot examples, as in Table~\ref{box:information_extraction_few_shot_prompt}, which provides examples of correctly extracted information from the speeches of different identities and skills, to improve the accuracy of the task as well as to align it with the type of output we expect.

\item Current information: Finally, we input the current speech of the player, the game state, e.g., the speaker's \emph{Player id}, role, the current speech types, as in Table~\ref{box:nlu_overall_prompt}, to prompt LLMs for deductive reasoning.

\end{itemize}

\label{sec:llm_prompt_for_nlg}
The speech generation prompt for the \Presenter module contains the following parts, as shown in Table~\ref{box:speech_generation_prompt}: 
\begin{itemize}[leftmargin=*]
\setlength{\itemsep}{0pt}
\item Description of the background of the Werewolf game, which is the same as in the \Listener module.

\item  (Optional) speech instruction.
The prompt is a structured output from the \Thinker module, and its meaning aligns with that of the \Listener module, with a 1-shot example.

\item  Task requirements, which is the similar to that in the \Listener module expect for the speech generation task.

\item  Current information, which is the similar to that in the \Listener module except that we prompt all the historical speeches.

\end{itemize}

\subsection{Game Log Examples}
Table~\ref{box:werewolf_game_log_example} presents a comprehensive analysis of a 9-player werewolf game log, culminating in a victory for the Werewolf.

\newpage

% part 1
\begin{table*}[!h]
\centering
\begin{tcolorbox}[
    colback=lightgray,    % 背景颜色
    % colframe=red!75!black,  % 边框颜色
    % left=5pt,                 % 左边距
    boxrule=1pt,              % 边框宽度
    % width=17cm,              % box 宽度
    title=Werewolf Game Background Prompt,
    fonttitle = \ttfamily,
]
{\ttfamily\small
\textbf{\# Task Scenario: 9-player Werewolf game speech.} \\

\textbf{"Good" Faction:} \\
- 3 Villagers \\
- 1 Seer \\
- 1 Witch \\
- 1 Hunter \\

\textbf{Werewolf Faction:} \\
- 3 Werewolves \\

\textbf{Common terminologies are explained as follows:} \\
1. Werewolf, bandit, wolf, bad faction, knife: Werewolf. \\
2. Villager, civilian, white card: Villager. \\
3. Seer, prophet: Seer. \\
4. Witch, witch card: Witch. \\
5. Hunter, gun: Hunter. \\
6. Gold, gold water, verified Good: A good person verified by the Seer. \\
7. Verify Kill: A Werewolf verified by the Seer. \\
8. Silver, silver water, Werewolves' target, Saved: A person saved by the Witch. \\
9. Iron, steel, certain: Very certain, e.g., "Player 3 is an iron Werewolf" or "Player 3 is definitely the Werewolf," indicates that Player 3 is certainly a Werewolf. \\
10. Jump: A player declares his/her role (not necessarily his/her true role). \\
11. Backstab: A Werewolf sides with the good people, betraying their own teammates. \\
12. Defame: To demean the identity of other players. \\
13. Exalt: To believe in the identity of other players. \\
14. Vote out, point, nominate, ballot: Voting, e.g., "Vote for Player 6 or Player 7," means to vote Player 6 or Player 7 out.
}
\end{tcolorbox}
\caption{Werewolf game background prompt.}
\label{box:werewolf_game_background_prompt}
\end{table*}

% part 2
\begin{table*}[!t]
\centering
\begin{tcolorbox}[
    % colback=lightgray,    % 背景颜色
    colback=lightgray,    % 背景颜色
    % colframe=red!75!black,  % 边框颜色
    % left=5pt,                 % 左边距
    boxrule=1pt,              % 边框宽度
    % width=17cm,              % box 宽度
    title=Speech Understanding Requirements Prompt,
    fonttitle = \ttfamily,
]
{\ttfamily\small
\textbf{\# Task requirements are as follows:} \\

Based on your understanding of the game state and speeches, please output the extraction results in JSON format in sequence. The format should be:
\begin{verbatim}
{
    "identities": {"<identity>": [player,player,...]} ,
    "actions": {"<action>": [subject player -> object player,
                subject player -> object player]}
}
\end{verbatim}

\textbf{\\Example:} 
\begin{verbatim}
{
    "identities": {"werewolf":[3,5]}, {"<action>": [subject player -> object player, 
                    subject player -> object player]}, }
    "actions": {"check":[1->6, 2->3]}
}
\end{verbatim}

- This indicates Players 3 and 5 are werewolves, Player 1 checks Player 6, and Player 2 checks Player 3.

- Player numbers can only be: 1, 2, 3, 4, 5, 6, 7, 8, 9. 

- When players express their intentions, please correspond to the identity of the player, for example, if Player 5 speaks, then consider from the perspective of Player 5.

- The subject number should be inferred from the context, such as 'I', 'you', 'he', 'she', etc. If unknown, use 'unknown', for example: "check":[unknown->6]. \\

\textbf{Possible JSON KEYs are:} \\

\textbf{Identities:}

- Roles: Seer, Witch, Hunter, Villager, Werewolf, "Good" faction, Werewolf faction, gold water, silver water, the werewolves' target, etc.

- Guess: suspicious, credible, uncertain, tolerant, etc.

- Speech: good (up), bad (down), listen well, listen to kill, etc.

- Faction: allied, support, werewolf candidate, etc.

- Online status: disconnected, offline, not online, voice, etc. \\

\textbf{Actions:}

\phantom{-} - Skills: \\ 
\phantom{-----} - Seer: check, inspect. \\
\phantom{-----} - Witch: poison, save. \\
\phantom{-----} - Hunter: shoot, take away, crash, kill. \\
\phantom{-----} - Werewolf: self-destruct, explode. 

\phantom{-} - And skills that will be used in the future: \\
\phantom{-----} - Vote: vote out, choose a target, etc. \\
        
- Quotes from other Players' statements do not need to be summarized.

- Note the distinction between quantifiers and player numbers: must be, that there are three werewolves.

- Note negative statements: not, impossible, implausible, not quite, etc.

- Note the abbreviation of number + information, e.g., "three golds, nine slashes, one, six, eight, three wolves" results in:{"identities": {"gold water":[3], "slash":[9], "werewolf":[1,6,8]}}
}
\end{tcolorbox}
\caption{Speech understanding requirements prompt.}
\label{box:dialogue_understanding_requirements_prompt}
\end{table*}

% part 3
\begin{table*}[!t]
\centering
\begin{tcolorbox}[
    colback=lightgray,    % 背景颜色
    % colframe=red!75!black,  % 边框颜色
    % left=5pt,                 % 左边距
    boxrule=1pt,              % 边框宽度
    % width=17cm,              % box 宽度
    title=Information Extraction Few-Shot Prompt,
    fonttitle = \ttfamily,
]
{\ttfamily\small
\textbf{\# The following are 11 speeches and corresponding information extraction examples:} \\

\textbf{Player 3 spoke:} 
"I checked Player 6, and I suggest Player 8 turn around and vote for Player 6. I will check the identity of Player 4 in the next round."

\begin{verbatim}
{
    "identities":{"seer":[3],"werewolf":[1,6,8]},
    "actions":{"check":[3->6],"suggest to vote":[8->6],
               "check in the next round":[3->4]}
}
\end{verbatim} \\

\textbf{Player 7 spoke:} 
"Player 2 and I are collaboratively searching for a Seer. Player 2 assists the good faction in combating werewolves. There's a possibility that Player 9 is a werewolf, although I am not certain. The behavior of Player 9 seems suspiciously similar to that of Player 2, who possesses the ability to shoot. Additionally, Player 4 is identified as a Witch. Regarding the usage of silver water, I suggest targeting Player 6."

\begin{verbatim}
{
    "identities":{"maybe a wolf":[9],"hunter":[2],"silver water":[4]},
    "actions":{"suggest to vote":[7->6]}
}
\end{verbatim} \\

\textbf{Player 9 spoke:} 
"Player 8 is the gold water. Player 2 is not a werewolf, neither is Player 3. However, Player 7 is suspicious, and I recommend voting against Player 7. The roles of Player 4 and Player 5 are unclear, and Player 1 suspects both of them to be werewolves. I advise Player 7 to use poison, which could help confirm my role as a Seer. Concerning the hunter, there is a standoff between Player 8 and myself. If there is any uncertainty about Players 1, 2, or 4, the gun should be used in this situation against Player 2. Now, it's time for Players 4 and 7 to present their arguments, and there is no need to focus on Player 9."

\begin{verbatim}
{
    "identities":{"gold water":[8],"good camp":[2,3],"suspicious":[7],
                  "werewolf":[4,5],"seer":[9] ,"werewolf candidate":[1,2,4],
                  "hunter":[2],"debate players":[4,7]},
    "actions":{"suggest to vote":[9->7],"suggest to poison":[unknown->7]}
}
\end{verbatim} \\

\textbf{Player 3 spoke:} 
"Being the first player to speak, my turn was conveniently arranged. However, I am uncertain about Player 2's allegiance. In my view, Player 2 lacks credibility."

\begin{verbatim}
{
    "identities":{"no result": []},
    "actions":{"no result": []}
}
\end{verbatim} \\

\textbf{Player 7 spoke:} 
"Player 3 will be poisoned tonight. I hold the Witch card. I heed the guidance of the two players with gold cards. Players 9 and 5 are identified as wolves. Players 4 and 6 hold cards corresponding to their numbers, with Player 4 being more trustworthy than Player 5. Player 3 cannot be revived. To preserve my own safety, I will reveal myself as the Witch. I have already used the silver water card on Player 1. Player 9 remarked that I should be pleased with this misfortune, indicating that the prime werewolf card was passed to a fellow teammate."

\begin{verbatim}
{
    "identities":{"witch":[7],"gold water":[2],"werewolf":[9,5],"suspicious":[4]},
    "actions":{"suggest to poison":[7->3],"believe to be a silver water":[7->1]}
}
\end{verbatim} \\

\textbf{Player 8 spoke:} 
"Player 5 appears highly suspicious. He could either be a werewolf or might be deceiving his teammates. His failure to set wolf traps, dishonesty about the wheat sequence, and excessive talking during the first microphone turn is concerning. Players 6 and 7 might be superficial wolves. Player 7, however, seems to have a sensible perspective and could be part of the good camp. I recommend voting against Player 5."
}
\end{tcolorbox}
% \caption{Information Extraction Few-Shot Prompt}
% \label{box:information_extraction_few_shot_prompt}
\end{table*}

\begin{table*}[!h]
\centering
\begin{tcolorbox}[
    colback=lightgray,    % 背景颜色
    % colframe=red!75!black,  % 边框颜色
    % left=5pt,                 % 左边距
    boxrule=1pt,              % 边框宽度
    % width=17cm,              % box 宽度
    % title=Information Extraction Few-Shot Prompt,
    fonttitle = \ttfamily,
]
{\ttfamily\small
\begin{verbatim}
{
    "identities":{"suspicious":[5],"werewolf":[6,7],"good camp":[7]},
    "actions":{"suggest to vote":[8->5]}
}
\end{verbatim} \\

\textbf{Player 2 spoke:} 
"Regarding the game, my suspicion falls on Players 1, 5, 7, and 3 as potential wolves. The accusation by Player 3, however, is incorrect. I find Player 3's judgment flawed. It's frustrating. Similarly, I suspect that Players 1, 5, 7, and 3 are wolves according to Player 5's perspective. Let's test this theory. I propose we eliminate Player 5 today, and then I, as a Witch, will poison Player 7 tomorrow night. Observe the game's progression tomorrow, and you will see that both Player 5 and I, as Witches, agree on Player 2, and our views align with Player 3's decision. Therefore, I request that we focus on Player 5 first."

\begin{verbatim}
{
    "identities":{"werewolves' target":[3],"werewolf":[1,5,7],"witch":[2]},
    "actions":{"suggest to vote":[2->5, 2->7]}
}
\end{verbatim} \\

\textbf{Player 1 spoke:} 
"Player 6 is engaging in killing actions. Players 5 and 7 have been poisoned. Players 4 and 5 are both targeting Player 1. Player 3 has been stabbed, and it's possible that Players 2, 4, and 9 each represent a threat, akin to three knives. Player 5 has revealed themselves as the Witch and has provided Player 3 with a dose of silver water."

\begin{verbatim}
{
    "identities":{"seer":[1],"poison":[5,7],"depreciate":[4,5],
                  "werewolves' target":[3],"werewolf":[2, 4,9],"witch":[5]},
    "actions":{"check":[1->6],"believe to be a silver water":[5->3]}
}
\end{verbatim} \\

\textbf{Player 1 spoke:}
"I, Player 1, am part of the good faction. The focus of today's game is on Players 3 and 5. Player 9 might be a werewolf. I did not use any poison last night."

\begin{verbatim}
{
    "identities":{"good camp":[1],"werewolf":[9]},
    "actions":{"suggest to vote":[1->3,1->5]}
} 
\end{verbatim} \\

\textbf{Player 9 spoke:}
"I am the Hunter. Player 7 has self-destructed. Player 2 might be associated with the silver water. As for myself, I reiterate that I am the Hunter. Player 1 is acting suspiciously, resembling a white card. I request the Witch to acknowledge this. Player 3 is overly concerned with external cards, which is uncharacteristic of a Prophet. Players 3 and 8, please return to the game, as there's still an opportunity for a round of confrontation."

\begin{verbatim}
{
    "identities":{"hunter":[9],"self-destruction":[7],"silver water and seer":[2],
                  "white":[1],"not like a seer":[3] },
    "actions":{"suggest to vote":[9->3,9->8]}
}
\end{verbatim} \\

\textbf{Player 4 spoke:} 
"I believe Player 6 is trustworthy as he revealed Player 6's key card. My intention is to verify Player 3. Player 7, who holds the gold water, should cast their vote against Player 8. It's evident that Players 3 and 7 are not the same individual. On the field, there are only two players acting as villagers. I have identified the three wolves. There is no necessity to doubt Player 7; instead, Player 4 can be acknowledged as the Seer."

\begin{verbatim}
{
    "identities":{"gold water":[7],"seer":[4]},
    "actions":{"consider credible":[4->6],"verified":[4->3],
               "suggest to vote":[4->8]}
}
\end{verbatim} 
}
\end{tcolorbox}
\caption{Information extraction few-shot prompt.}
\label{box:information_extraction_few_shot_prompt}
\end{table*}

% part 4
\begin{table*}[!h]
\centering
\begin{tcolorbox}[
    % colback=lightgray,    % 背景颜色
    colback=lightgray,    % 背景颜色
    % colframe=red!75!black,  % 边框颜色
    % left=5pt,                 % 左边距
    boxrule=1pt,              % 边框宽度
    % width=17cm,              % box 宽度
    title=LLM prompting for the \Listener,
    fonttitle = \ttfamily,
]
{\ttfamily\small
\textbf{\# Task type: Information Extraction} \\

\hyperref[box:werewolf_game_background_prompt]{\$\{\{ Werewolf Game Background Prompt \}\}} 

\hyperref[box:speech_generation_prompt]{\$\{\{ Speech Understanding Requirements Prompt \}\}}

\hyperref[box:information_extraction_few_shot_prompt]{\$\{\{ Information Extraction Few-Shot Prompt \}\}}

\textbf{\# The task text is as follows:} \\

Player 8 spoke: "I think Player 9 is a good person, but I am not sure about the identities of Player 5 and Player 6." \\

Please directly output the information extraction result in JSON format:

}
\end{tcolorbox}
\caption{LLM prompting for the \Listener.}
\label{box:nlu_overall_prompt}
\end{table*}

% nlg
\begin{table*}[!h]
\centering
\begin{tcolorbox}[
    % colback=lightgray,    % 背景颜色
    colback=lightgray,    % 背景颜色
    % colframe=red!75!black,  % 边框颜色
    % left=5pt,                 % 左边距
    boxrule=1pt,              % 边框宽度
    % width=17cm,              % box 宽度
    title=Speech Generation Prompt,
    fonttitle = \ttfamily,
]
{\ttfamily\small
Now that you play as a Werewolf player, I'm going to provide you with some information about the position you're about to speak in, which hasn't happened yet and is not historical information, and ask you to concatenate this information to generate a paragraph of speech text. \\

First, I'll give you some background on the game: \\

\textbf{\# Task type: Game Dialog Generation} \\

\hyperref[box:werewolf_game_background_prompt]{\$\{\{ Werewolf Game Background Prompt \}\}} 

You are playing a 9-player werewolf game. Suppose you're game Player 1, and your identity is Seer. \\

I provide you with the format of the in-field message:
\begin{verbatim}
{
    "identities": {"<identity>": [player,player,...]} ,
    "actions": {"<action>": [[subject player, object player], 
                [subject player, object player]]}
}  
\end{verbatim} \\

\textbf{Example:}
\begin{verbatim}
{
    "identities": {"werewolf":[3,5]}
    "actions": {"check":[[1,6],[2,3]]}
}
\end{verbatim} \\

- Indicate that Player 3 and Player 5 are werewolves, Player 1 checks Player 6, Player 2 checks Player 3, and the subject and object are irreversible. 

- The only possible player IDs are 1,2,3,4,5,6,7,8,9, and unknown should be replaced by the speaker's player ID. \\

Note that the generated speech result should strictly fulfill the following 10 requirements:

1. Include all the information in the information extraction result.

2. Don't over-imagine and introduce hallucination, and prioritize the accuracy of the information.

3. The logic between the generated results should be in line with the position of the players in Wolfsbane, and there should not be any contradictions between the logic before and after.

4. Pay attention to the diversity of generated results.

5. The generated results should be as anthropomorphic as possible, imitating the speaking style of human players.

6. Please be firm in your belief that you are the Good faction, whether you yourself are in the Good faction or the Werewolf faction.

7. Identities or actions can be left out if the result is empty, empty is invalid information.

8. A player can only be one of the roles of Villager, Seer, Witch, Hunter, or Werewolf, for example, it's impossible to be a Witch and a Hunter at the same time, if there is more than one conflicting Werewolf identity in the information I've provided you with, please randomly choose one.

9. Please state your identity first.

10. Do not make statements such as "unite", "stay alert", "defend the village", "together we will go to victory", "find out the werewolf as soon as possible", "keep watching", "the information shows up", "hopefully we will find it", "think about what we should do next", "keep discussing" or similar statements. \\
}
\end{tcolorbox}
% \caption{Speech Generation Prompt}
% \label{box:speech_generation_prompt}
\end{table*}

\begin{table*}[!t]
\centering
\begin{tcolorbox}[
    % colback=lightgray,    % 背景颜色
    colback=lightgray,    % 背景颜色
    % colframe=red!75!black,  % 边框颜色
    % left=5pt,                 % 左边距
    boxrule=1pt,              % 边框宽度
    % width=17cm,              % box 宽度
    % title=Speech Generation Prompt,
    fonttitle = \ttfamily,
]

{\ttfamily\small
Now that you are Werewolf Player 1, with the identity of Seer, I am providing you with the key information for your upcoming statement:

\begin{verbatim}
{
    "identities": {"villager": [1]}, 
    "actions": {"no result": []}
} 
\end{verbatim} \\

Please follow the key messages to the letter and keep the text under 100 words:

\begin{verbatim}
{
    "identities": {"villager": [1]}, 
    "actions": {"no result": []}
}
\end{verbatim} \\

Now it's your turn to speak: 

}
\end{tcolorbox}
\caption{Speech generation prompt.}
\label{box:speech_generation_prompt}

\end{table*}

% Case Study 样式需要修改
\begin{table*}[!h]
\centering
\begin{tcolorbox}[
    colback=lightgray,    % 背景颜色
    % colframe=NavyBlue,  % 边框颜色
    % left=5pt,                 % 左边距
    boxrule=1pt,              % 边框宽度
    % width=17cm,              % box 宽度
    title=Werewolf Game Log Example,
    fonttitle = \ttfamily,
]
{\ttfamily\small
\textbf{game role assignment:} \\
- Player 1: Werewolf. \\
- Player 2: Witch. \\
- Player 3: Villager. \\
- Player 4: Werewolf. \\
- Player 5: Villager. \\
- Player 6: Hunter. \\
- Player 7: Villager. \\
- Player 8: Werewolf. \\
- Player 9: Werewolf. \\

\textbf{day 1, night:} \\
- Werewolves: Player 4, Player 8, and Player 9 choose to kill Player 5. \\
- Seer: Player 1 investigates Player 7, who is in the good camp. \\
- Witch: the werewolf's target is Player 5, and the witch chooses to use a potion to save Player 5. \\

\textbf{day 1, announcement:} \\
- no players were out last night. \\
- remaining players: Player 1(Werewolf), Player 2(Witch), Player 3(Villager), Player 4(Werewolf), Player 5(Villager), Player 6(Hunter), Player 7(Villager), Player 8(Werewolf), Player 9(Werewolf). \\

\textbf{day 1, discussion: } \\
- (random, Player 9 starts to speak) \\
- day 1 Player 9 (Werewolf) said: "Good man card, ah, the first position to speak is not my specialty, you say the system and gave me this right, so that I can openly paddling, hey, I especially want to exercise, I will say Player 2, just more than a sentence do not say it, OK? On the first day to find the next day we in that what in the look at the side of the station to update the speech, okay? Player 9 is a white ah, Player 9 Player 8 license plate" \\
- day 1 Player 1 (Seer) said: "The Player 7 is gold water, with a sense of hearing, to lead the good guys to victory. My last words are to keep an eye on Player 8 and Player 9 as they are close friends." \\
- day 1 Player 2 (Witch) said: "Player 7 has an average sense of hearing and cannot take sides. Player 1 listens to the situation and listens to the speech. Player 7 only listens to Player 8 or Player 9, do the other cards enter the perspective? Player 5 is a silver water cutter, I saved him, don't hold a grudge. Before Player 6, Player7 and Player 8 werewolves should not be less, can go to the highest identity." \\
- day 1 Player 3 (Villager) said: "I think Player 4 need be intimidated, Player 7 has no sincerity, Player 9's speech is error tolerant, Player 2 keeps taking the Witch card, and he's good." \\
- day 1 Player 4 (Werewolf) said: "I'll give gold water on Player 2, and Player 9 is labeled as a werewolf. Player 7 is a predecessor of the Player 5 and Player 2 werewolves. Player 3 will not be a werewolf. Witch card Player 2 doesn't report silver water, which is kind of cool. I'll go ahead and vote for Player 1, with a high probability that Player 9 is a werewolf." \\
- day 1 Player 5 (Villager) said: "Player 4 poison Player 1, no need for a Seer. Player 5 has a sense of hearing, Player 7 just doesn't fight because he respects his predecessor. Player 2 is not a one-burst, Player 9 must be saved. Convinced by virtue, vote Player 4 poison Player 1 or Player 2 are marked." 
}
\end{tcolorbox}
% % \caption{Werewolf Game Log Example}
% \label{box:werewolf_game_log_example}
\end{table*}

\begin{table*}[!h]
\centering
\begin{tcolorbox}[
    colback=lightgray,    % 背景颜色
    % colframe=red!75!black,  % 边框颜色
    % left=5pt,                 % 左边距
    boxrule=1pt,              % 边框宽度
    % width=17cm,              % box 宽度
]
{\ttfamily\small
- day 1 Player 6 (Hunter) said: "Very well, great, Player 4 feel the back side does not need to talk about what too much, Player 2 Seer do not want it, right? If it's a werewolf, I'm sure I'd want neither." \\
- day 1 Player 7 (Villager) said: "Deacon Player 3 is a white, Player 6 is a top werewolf, Player 4, Player 6, Player 8 three Werewolves, Player 1, Player 6, Player 9 are numbered cards, Player 4 is a werewolf, Player 7 suggests betting on Player 6, and the ticket dies poison Player 1." \\
- day 1 Player 8 (Werewolf) said: "Player 8 questions Player 1's board and thinks he has intentions. Player 1's statement mentions checking Player 7, but it doesn't sound good. Player 4's statement is off and suggests washing the oil and playing PK table water tomorrow." \\

\textbf{day 1, voting: } \\
- voting information: 1 voted for 1, 2 voted for 1, 3 voted for 1, player 4 abstained, 5 voted for 4, 6 voted for 4, 7 voted for 4, 8 voted for 4, 9 voted for 4 \\
- voting result: Player 4 \\
- remaining players: Player 1(Werewolf), Player 2(Witch), Player 3(Villager), Player 5(Villager), Player 6(Hunter), Player 7(Villager), Player 8(Werewolf), Player 9(Werewolf). \\

\textbf{day 1, last words:} \\
Player 4 (Werewolf) Last Words: "We should vote Player 5 out and poison one werewolf, I don't care if I vote them all out. Player 1 poisoned Player 2, Player 2 witch tags down to find the Hunter. The werewolves may kill Player 2 at night, the gun card should hide. Witch card not down tomorrow, you lead the team down, gun card hidden knife hook." \\

\textbf{day 2, night:} \\
- Werewolves: Player 4, Player 8, and Player 9 choose to kill Player 2. \\
- Seer: Player 1 investigates Player 3, who is in the good camp. \\
- Witch: Player 2 poisoned Player 1. \\

\textbf{day 2, announcement:} \\
- players who died last night: Player 1, Player 2 \\
- remaining players: Player 3(Villager), Player 5(Villager), Player 6(Hunter), Player 7(Villager), Player 8(Werewolf), Player 9(Werewolf). \\

\textbf{day 2, discussion: } \\
- start from the right of the dead player, Player 3 starts to speak) \\
- day 2 Player 3 (Villager) said: "Player 4 is a true pre-card, Player 5 offers to vote Player 4 poison Player 1, 5, 6, 7 and 8, there is no opposition, Player7 also has a familiar moderation to him, Player 4'attitude is there, it is not difficult to find a prophet. Player 6 played Player 4, 6 and 8 in the werewolf pit, Player 6 and Player 7 played awkwardly in the werewolf pit. I think Player 4 is the true Seer, and Player 1 is a werewolf card that goes to silver water. Vote Player 4 today, and Player 7 says vote Player 4 and Player 8, where is team Player 4 rolling?" \\
- day 2 Player 5 (Villagers) said: "Right ah, you can hear out the Player 4 times the Seer, and other people can not hear out ah, on this chat he wrote to write he went out what is the problem ah? Eight some words should not be put on this field to say the end, you can go to say well, seven or eight shun down well, hey, over." \\
- day 2 Player 6 (Hunter) said: "The Seer points, the next seven is finished. Seven doesn't know what the card is and is completely unintelligible. Ticket four poison one poison six dry six. Card seven must be a werewolf card." \\
- day 2 Player 7 (Villager) said: "Seven is a polearm sign, four is probably a wolf pre, and six is probably a werewolf. Ticket seven, but I would go for ticket six. There is a white card and a werewolf card in eight or nine." 
}
\end{tcolorbox}
% \caption{Werewolf Game Background Prompt}
% \label{}
\end{table*}

\begin{table*}[!h]
\centering
\begin{tcolorbox}[
    colback=lightgray,    % 背景颜色
    % colframe=red!75!black,  % 边框颜色
    % left=5pt,                 % 左边距
    boxrule=1pt,              % 边框宽度
    % width=17cm,              % box 宽度
]
{\ttfamily\small
- day 2 Player 8 (Werewolf) said: "Player 8 suggests racking silver water and hammering Seven for a deep push on Werewolf. Player 6 and Player 9 are two werewolves' targets, and Player 6 is a werewolf. Player 1 and Player 7 are two werewolves, Player 7 could be the werewolf. Player 9 spoke first, two white cards in Player 7 and Player 8 in the back seat, and all dumped gold in the front seat. Player 6 is the most bandit to speak, and is killed by listening. It is recommended that Player 6 be ordered off first." \\
- day 2 Player 9 (Werewolf) said: "Player 7 and Player 8 are a double to play Player 6. Player 5 has returned to the ticket, and the Player 7 and Player 8 goes down to a point Player 7. Player 6 is a white card, Player 7 and Player 8 are sold." \\

\textbf{day 2, voting: } \\
- voting information: 3 votes for 7, 5 votes for 7, 6 votes for 7, 7 votes for 6, 8 votes for 6, 9 votes for 7 \\
- voting result: Player 7 \\
- remaining players: Player 3(Villager), Player 5(Villager), Player 6(Hunter), Player 8(Werewolf), Player 9(Werewolf). \\

\textbf{day 2, last words:} \\
Player 7 (Villager) Last Words: "Vote out or poison Player 4, Player 1 has the same perspective as Player 5, and what was dished out on Player 6? Player 3 may be a werewolf. Player 2 is a werewolf, leaving a three-in-two pattern inside the Player 8 and Player 9, and placing Player 3. Player 9 could be a werewolf."\\

\textbf{day 3, night:} \\
- Werewolves: Player 8, and Player 9 choose to kill Player 5. \\
- Seer: the Seer is out. \\
- Witch: the Witch is out. \\

\textbf{day 3, announcement:} \\
- players who died last night: Player 5 \\
- remaining players: Player 3(Villager), Player 6(Hunter), Player 8(Werewolf), Player 9(Werewolf). \\

\textbf{day 3, discussion: } \\
- start from the left of the dead player, Player 3 starts to speak) \\
- day 3 Player 3 (Villager) said: "I do not know whether the double wolf has control of the field ah, we are empty, randomly on the three ends of the inside point it ah, there is no empty field on the point of eight, because I suddenly thought of this nine last speech is too brief." \\
- day 3 Player 9 (Werewolf) said: \\
\phantom{-} - policy intent: 
\begin{verbatim}
{
    "identities": {"werewolf": [8,9]},
    "actions": {"suggest to vote": [[9,6]]}
}
\end{verbatim}

\phantom{-} - [\textbf{GPT3.5-LtM}]: "Hello everyone, I'm Player 9 and I'm a civilian. After the first two days of the game, we have managed to identify a werewolf and the two good guys are out. Now that our identities have been revealed, the only people left in the werewolf camp are me and Player 8. Based on the speeches and votes from the first two days of play, I can confirm that Player 6 is the good guy because he was voted to death by Player 1, who was the good guy who was knifed by us. Therefore, I suggest we return the vote to Player 6 and vote him out. We can use this vote result to increase our chances of finding the werewolf, as well as reduce casualties in our good guy camp. I hope you can understand and support my proposal. Thank you!" \\
}

\end{tcolorbox}
% \caption{Werewolf Game Background Prompt}
% \label{}
\end{table*}

\begin{table*}[t]
\centering
\begin{tcolorbox}[
    colback=lightgray,    % 背景颜色
    % colframe=red!75!black,  % 边框颜色
    % left=5pt,                 % 左边距
    boxrule=1pt,              % 边框宽度
    % width=17cm,              % box 宽度
]
{\ttfamily\small
\phantom{-} - [\textbf{GPT3.5-T}]: "Hello, fellow villagers. As a villager, I have been actively involved in the game discussions, but unfortunately, we have yet to find the werewolf. I'd like to start by reviewing the previous two days' speeches, and we've found a few clues. First of all, based on yesterday's vote, Player 4 was voted out of the game by the referendum, and in his last words, he mentioned the vote Player 5 poison and one wolf, which may imply that he had some important information. Also, according to last night's deaths, Player 2 was knifed by a werewolf, which means our hunter is out. For this, I express my deep regret." \\

\phantom{-} - [\textbf{GPT4-LtM}]: "As Werewolf Player 9, I'm going to firmly consider myself a good guy. Hello everyone, I am Player 9 and I am a good guy. From the previous two days of voting, I voted for Player 6 on the second day, and that was my personal judgment. Today I would like to say that I suspect Player 8 and I are on the same side. These are some of my judgments on the field, and I hope you will consider them carefully." \\

\phantom{-} - [\textbf{Finetune-T}]: "I'm Player 9, and I think this Player 6 is too good to just vote him out. I now have Player 8 wolf cards, just don't get out. These wolf cards are so powerful, no one can beat them." \\

\phantom{-} - [\textbf{GPT4-T}]: "I'm Player 9, and I'm a werewolf. I will say that Seven, who everyone voted for yesterday, is actually a hunter, and I understand Player 6's questioning of him, but I personally feel that the logic of Player 6's statement was a bit messed up, and I'm a bit skeptical of him. Player 8 offered to hammer Player 7 yesterday, and I'm all for his decision, but I feel like he's extrapolating a little too arbitrarily on Player 6 and Player 9. Also, Player 3, you accused me of being brief, and I'd like to point out that that doesn't make me a werewolf. So my suggestion is that we should address Player 6 first and then look at Player 8." \\

- day 3 Player 8 (Werewolf) said: "Can't kill Player 8 anymore, Player 8 is a werewolf come gun hang yourself." \\
- day 3 Player 6 (Hunter) said: "Player 6 is a Werewolf, step on." \\

\textbf{day 3, voting: } \\
- voting information: 3 abstained, 6 voted for 6, 8 abstained, 9 voted for 6. \\
- voting result: Player 6 \\

\textbf{game result: } \\
the Werewolves win the game. (final surviving player: Player 3 (Villager), Player 8 (Werewolf), Player 9 (Werewolf))

}
\end{tcolorbox}
\caption{Werewolf game log example.}
\label{box:werewolf_game_log_example}
\end{table*}

\twocolumn
\end{document}